%% file: elsarticle-MUSE.tex
\documentclass[authoryear,nopreprintline,review,12pt]{elsarticle}

\usepackage{amssymb}
\usepackage{amsmath}
\usepackage{amsthm}

\input{texcommands}

\usepackage{dsfont}
\usepackage{multicol}
\usepackage{xcolor}
\usepackage{parskip}
\usepackage{color,soul}
\usepackage{subcaption}
\usepackage{caption}
\usepackage{stackengine}
\usepackage[utf8]{inputenc} 
\usepackage[T1]{fontenc}    
\usepackage{hyperref}       
\usepackage{url}            
\usepackage{booktabs}       
\usepackage{nicefrac}       
\usepackage{wrapfig}
\usepackage[none]{hyphenat}
\usepackage{longtable}
\usepackage[tableposition=below]{caption}
\usepackage{enumitem}
\usepackage{natbib}
\usepackage{graphicx}
\usepackage{xcolor}
\usepackage{tikz}
\usepackage{pifont}
\usepackage{setspace}
\usepackage{algorithm}
\usepackage{algpseudocode}

\definecolor{green}{RGB}{36, 214, 36}
\definecolor{red}{RGB}{235, 30, 30}
\definecolor{lightredshade}{HTML}{dea9a9}
\definecolor{lightgreenshade}{HTML}{bce3bd}
\definecolor{lightblueshade}{HTML}{cacbe8}
\definecolor{MyDarkBlue}{rgb}{0,0.08,1}
\definecolor{MyDarkGreen}{rgb}{0.02,0.6,0.02}
\definecolor{MyDarkRed}{rgb}{0.8,0.02,0.02}
\definecolor{MyDarkOrange}{rgb}{0.40,0.2,0.02}
\definecolor{MyPurple}{RGB}{111,0,255}
\definecolor{MyRed}{rgb}{1.0,0.0,0.0}
\definecolor{MyGold}{rgb}{0.75,0.6,0.12}
\definecolor{MyDarkgray}{rgb}{0.66, 0.66, 0.66}

\definecolor{MyYellow}{rgb}{254, 246, 170}
\definecolor{MyBlue}{rgb}{170, 217, 251}

\theoremstyle{definition}

\usepackage{enumitem}

\begin{document}

\begin{frontmatter}



\title{Competence-Aware AI Agents with Metacognition for Unknown Situations and Environments (MUSE)}


\author[label1]{Rodolfo Valiente} 
\author[label1]{Praveen K. Pilly} 

\affiliation[label1]{organization={Intelligent Systems Center, HRL Laboratories},
            addressline={3011 Malibu Canyon Road}, 
            city={Malibu},
            postcode={90265}, 
            state={CA},
            country={USA}}

\begin{abstract}
Metacognition, defined as the awareness and regulation of one's cognitive processes, is central to human adaptability in unknown situations. In contrast, current autonomous agents often struggle in novel environments due to their limited capacity for adaptation. We hypothesize that metacognition is a critical missing ingredient in autonomous agents for the cognitive flexibility needed to tackle unfamiliar challenges. Given the broad scope of metacognitive abilities, we focus on competence awareness and strategy selection. To this end, we propose the Metacognition for Unknown Situations and Environments (MUSE) framework to integrate metacognitive processes of self-assessment and self-regulation into autonomous agents. We present two implementations of MUSE: one based on world modeling and another leveraging large language models (LLMs). Our system continually learns to assess its competence on a given task and uses this self-assessment to guide iterative cycles of strategy selection. MUSE agents demonstrate high competence awareness and significant improvements in self-regulation for solving novel, out-of-distribution tasks more effectively compared to model-based reinforcement learning and purely prompt-based LLM agent approaches. This work highlights the promise of approaches inspired by cognitive and neural systems in enabling autonomous agents to adapt to new environments while mitigating the heavy reliance on extensive training data and large models for the current models.
\\
\end{abstract}



\end{frontmatter}


\section{Introduction}

The pursuit of fully autonomous agents in artificial intelligence (AI) remains a significant challenge. Current autonomous agents are primarily designed for operating environments, conditions, and uses that are known \emph{a priori}. They rely on either scripted behaviors or pre-trained policies, both of which struggle to handle unknown situations effectively. As a result, when faced with novelty, they are prone to fail with suboptimal or even catastrophic outcomes (e.g., robotic manipulation errors in unstructured settings). This limitation severely restricts their deployment in safety-critical unknown environments, especially for long-duration missions or applications with little to no human oversight. Therefore, there is a practical urgency to reduce the failure rate, time to completion, and cost of autonomous missions by enabling resilient handling of unknowns during deployment. 

In mainstream AI, large-scale multi-task pre-training has emerged as the leading approach for enhancing adaptability in autonomous agents ~\citep{openendedlearningteam2021openended}. For example, Adaptive Agent (AdA)~\citep{adaptiveagentteam2023humantimescale} was trained with billions of frames and tasks to enable rapid adaptation to unseen, open-ended tasks. Similarly, the RT-2~\citep{brohan2023rt2} and RT-X~\citep{padalkar2023open} models leverage large-scale robotic trajectory datasets to train agents capable of solving novel manipulation tasks and generalizing to new robots and environments. However, internet-scale pre-training to be able to handle every potential change and combination of changes in real-world applications is impractical, prohibitively resource-intensive, and expensive. Even with significantly limited data, AI agents must intelligently interpolate and extrapolate beyond their pre-trained scenarios while continually learning and adapting to novelty ~\citep{kudithipudi2023}. In other words, when faced with novel scenarios, pre-trained knowledge must be continually updated to strike a dynamic balance between stability and plasticity~\citep{grossberg1980}. 

Similar to humans, AI agents can leverage pre-deployment training to acquire a wide range of skills across diverse, known scenarios. Importantly, they can also be equipped with the ability to engage in online learning for continual improvement when encountering novel situations. For example, a teenager attending driving school follows a structured curriculum that teaches foundational vehicle control skills, which are then progressively built upon to master more complex tasks, such as merging onto highways or navigating construction zones. This learning process is cumulative in the sense that mastery of foundational skills simplifies the acquisition of more advanced ones. Moreover, the key principles of driving are consolidated in the student’s brain, protecting them from catastrophic forgetting. Ultimately, the end of driving school marks the beginning of a lifelong learning process, where the student draws on prior experiences to navigate novel driving challenges independently without an instructor. 

Metacognition, defined as the awareness and regulation of one's cognitive processes, is a key human trait that has been extensively studied in cognitive psychology~\citep{flavell1979, nelsonnarens1990, metcalfe1993cue, koriat1997monitoring, dunlosky2008metacognition}. This metacognitive flexibility enables humans to learn online efficiently and solve problems iteratively, especially in relation to new tasks. For instance, students can leverage metacognition to more accurately assess their knowledge and adjust their study habits accordingly ~\citep{cohen2012importance, chen2017strategic}. The use of metacognition among college students has indeed been shown to correlate significantly with various measures of academic success ~\citep{young2008metacognitive,isaacson2006metacognitive}. Students who perform poorly often overestimate their abilities, leading to under-preparation for exams. This is a common issue in education, where overconfident students allocate insufficient time to study, believing they have already mastered the material. Conversely, students who underestimate their knowledge may spend excessive time reviewing topics they already understand, hindering their progress. Even children as young as three years old can become effective learners and thinkers from activities designed to develop metacognitive skills ~\citep{chatzipanteli2014development}. It is generally agreed that self-assessment of competence, which can range from over-confidence in novices to slight under-confidence in experts~\citep{kruger1999,dunning2011dunning}, is a capability that is teachable and improves naturally as one becomes more skilled or knowledgeable~\citep{kramarski2003, schraw2006}. Neuroscience further reveals that the subregions of the prefrontal cortex responsible for metacognitive judgments are distinct from those involved in cognitive functions like visual memory recognition, such that the metacognitive function can be selectively deactivated without affecting the cognitive function (e.g.,~\citet{middlebrooks2012, miyamoto2017causal}).

\begin{figure}[h!]
    \centering
    \includegraphics[width=0.8\linewidth]{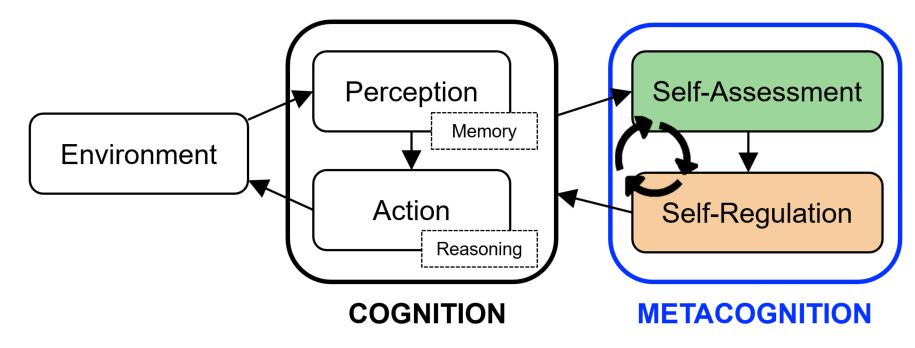}
    \caption{The metacognitive cycle of self-assessment and self-regulation operates on the traditional perception-action loop of existing AI agents to boost their ability for iterative problem-solving in unknown situations and environments.}
    \label{fig:metacognitive_cycle}
\end{figure}

While metacognition spans a wide range of capabilities and higher-order cognitive processes (e.g., Feeling of Knowing, Judgment of Learning, Source Monitoring), it can be conceptualized as an internal perception-action loop of self-assessment and self-regulation~\citep{nelsonnarens1990, dunlosky2013handbook}. Self-assessment in this context refers to an individual's ability to assess their competence regarding a specific task. Self-regulation refers to the ability to strategically select and control one's actions based on this self-assessment. In this article, we introduce the \textit{Metacognition for Unknown Situations and Environments} (MUSE) framework to computationally instantiate and train the metacognitive capabilities of self-assessment and self-regulation for AI agents, so that they can also achieve more efficient learning and improved generalization to unknown scenarios (Figure~\ref{fig:metacognitive_cycle}). Specifically, the self-assessment mechanism is designed to predict the agent's likelihood of successfully completing a given task for proposed action plans, with a learnable internal model informed by past experiences. And the self-regulation mechanism leverages this self-assessment to enable iterative cycles of competence-aware strategy selection for problem-solving. We present two implementations of the MUSE agent: one based on world modeling and the other utilizing large language models (LLMs). Our experiments in two distinct environments (namely, Meta-World and ALFWorld) demonstrate that MUSE agents achieve substantial improvements in handling novel scenarios compared to baseline/non-metacognitive approaches. Further, we show that metacognition makes a particularly big impact on smaller or less-capable LLM agents, making them amenable to edge deployment as well as less reliant on big data for online adaptation. 

In contrast to current reinforcement learning (RL) approaches (e.g., ~\citet{silver2016mastering, silver2017, silver2018, ha2018world, hafner2023mastering}), which focus on maximizing expected cumulative reward, our MUSE framework prioritizes competence as the primary evaluation metric to enhance the agent's adaptability to unknown situations. We hypothesize that focusing solely on maximizing return may cause an agent to become stuck in unfamiliar situations, particularly under sparse reward regimes. By contrast, maximizing competence continually for strategy selection encourages more effective exploration in such novel environments. Through metacognition, the agent can evaluate its capabilities and attempt new strategies within its perceived competence, enabling safer and more effective online adaptation. In other words, by maximizing competence, the agent not only improves its ability to tackle immediate challenges more effectively but also fosters iterative problem-solving in complex environments.

The contributions of our article are threefold:
\begin{itemize}
    \item We introduce the Metacognition for Unknown Situations and Environments (MUSE) framework, which integrates metacognitive functions of self-assessment and self-regulation into sequential decision-making agents.
    \item We propose an implementation of a competence awareness model that continually learns to assess the agent's competence on a given task and serves as an evaluation function for planning (Self-Assessment).
    \item We propose an implementation of a policy modulation model that leverages self-assessment to iteratively drive strategy selection by identifying action plans that maximize the likelihood of task success (Self-Regulation).
\end{itemize}

\section{Related Work}
\subsection{Self-Assessment}
Self-assessment broadly refers to an agent's ability to monitor its internal states and assess its capabilities and performance in relation to tasks and goals. It enables humans to reflect and make adjustments for improving outcomes~\citep{flavell1979, nelsonnarens1990, schraw1998promoting}.

\textbf{World Models}: These are generative models of environmental dynamics~\citep{ha2018world, robine2023transformer, micheli2023transformers, hansen2022modem, hansen2023td}, which can be used to estimate the expected cumulative reward of agents. These models often employ sequence-based architectures, such as recurrent neural networks, to predict the next state, reward, and terminal signals. Decoder-based World Models~\citep{hafner2023mastering, robine2023transformer, micheli2023transformers} can additionally generate the input state corresponding to the predicted next state. In contrast, decoder-free World Models focus on predicting the outcomes of actions in the latent space, bypassing the need to decode input states~\citep{hansen2022modem, hansen2023td}.
World Models act as proxy simulators during training, enabling agents to learn more efficiently by reducing the reliance on real-environment interactions (e.g., ~\citet{ha2018world, koul2020dream, hafner2023mastering}). In this work, we extend the capabilities of decoder-based World Models by training them to predict not only environmental dynamics but also the agent's competence to solve a given task.

\textbf{LLM Critics}: These models are designed to evaluate the performance of LLMs. The LLM itself can be prompted to provide feedback on its outputs~\citep{madaan2023self}, intermediate reasoning steps~\citep{paul2023refiner}, or even the prompt itself~\citep{hu2023evoke}. Some approaches enhance the correctness and quality of LLM outputs by using stochastic beam search guided by self-evaluation~\citep{xie2023decomposition}. Recognizing that LLMs currently are limited in identifying their own errors, reasoning missteps, or biases~\citep{huang2023large}, researchers have augmented LLM critics with external tools, such as search engines and calculators, to improve reliability~\citep{gou2023critic}. Retrieval Augmented Generation (RAG) approaches have also been proposed to strengthen self-evaluation by providing relevant external knowledge bases~\citep{asai2023self}. In contrast, MUSE does not rely solely on pre-trained knowledge or external tools. Instead, it continually learns and grounds itself to evaluate its competence on given tasks and uses this kind of self-assessment to modulate policy decisions. 

\textbf{Confidence Networks}: Recent work in cognitive computational neuroscience has developed quantitative frameworks for assessing metacognitive judgments related to self-assessment across a range of domains, task difficulties, and time scales with and without external feedback~\citep{fleming2024, lu2025}. Measures that have been proposed and utilized to assess metacognition include the statistical correlation between self-reported confidence ratings and actual performance across trials as well as the more reliable meta-\emph{d’} metric, which measures the ability of self-assessment to discriminate between high-performance (correct) and low-performance (incorrect) trials without being affected by response bias in metacognitive judgments~\citep{maniscalco2012, fleming2014}. 

Consistent with these measures of metacognition, there is prior work in machine learning aimed at self-assessment of deep neural networks that perform perception tasks. For classification, ~\citet{corbiere2019} trained a separate neural network (called ConfidNet) that operates on high-level features extracted by the classifier to predict the true class probability (TCP), which is the softmax probability of the correct class irrespective of whether it was chosen or not. Further, ~\citet{webb2023} trained confidence networks for a variety of perception tasks to instead predict the probability of the decision being correct, i.e., a value of 1 if correct and 0 otherwise. They also trained a RL agent that chooses among perception labels as well as an opt-out action that earns a low-risk low-reward in situations when the agent is least certain about its decision. These self-assessment metrics themselves are task-agnostic~\citep{fleming2024} but require a training scheme that is adapted to the specifics of the task. In this regard, one of our contributions is to implement a global self-assessment metric for artificial agents that predicts the probability of task success in each episode over time. 

\subsection{Self-Regulation}
Self-regulation is the process by which an agent dynamically adjusts its behavior based on self-assessment to achieve specific goals~\citep{flavell1979, nelsonnarens1990}. This ability is essential for humans to exhibit robust decision-making and function autonomously in unfamiliar environments. 

\textbf{Model-based Reinforcement Learning (MBRL) Agents}: These agents utilize World Models to simulate future scenarios, enabling them to train with minimal real interactions with the environment~\citep{moerland2023}. Dyna~\citep{sutton1991} is a foundational architecture that integrates learning and planning within a single agent. Its core idea is to use the agent's real experiences to update not only its policy but also its internal model of the environment, and thereby leverage the updated internal model to further improve the policy offline using simulated experiences. Among recent leading ones, AlphaGo~\citep{silver2016mastering}, AlphaZero~\citep{silver2017}, and MuZero~\citep{silver2018} are all MBRL systems that employ internal simulations/models and Monte Carlo Tree Search (MCTS) to explore potential action paths, based on variations of the Upper Confidence Bound (UCB) score, and evaluate them for action selection. In contrast to MBRL agents, which prioritize maximizing expected cumulative reward, the self-regulation mechanism of MUSE leverages competence as the primary evaluation metric to enhance the agent's ability to navigate and adapt effectively in unknown situations. 

\textbf{Prompt-based LLM Agents}: The capabilities of LLMs extend beyond language generation, making them increasingly popular for reasoning tasks to potentially deal with novelty. Chain-of-Thought (CoT) prompting~\citep{wei2022chain}, for example, decomposes a complex problem into intermediate steps to arrive at a final answer. However, CoT reasoning struggles to yield accurate results due to error propagation as the number of steps increases ~\citep{chen2022program}. Techniques such as self-consistency~\citep{wang2022self}, least-to-most prompting~\citep{zhou2022least}, and Tree-of-Thought (ToT) prompting~\citep{yao2024tree} aim to mitigate this issue by improving sampling strategies and leveraging search algorithms. Nevertheless, these methods rely solely on the LLM's pre-trained knowledge, which limits their ability to adapt to external feedback.
 
Beyond reasoning tasks, LLMs are also being increasingly applied to operate in an agentic loop of perception and action, which unlocks the benefits of large-scale pre-training for multi-step interactive tasks without relying on RL. ReAct~\citep{yao2023react} was among the first purely prompt-based LLM agents that integrated both reasoning and action planning to perform text-based problem-solving. However, ReAct is inefficient and limited in its ability to transfer performance improvements to subsequent episodes. To address this issue, Reflexion~\citep{shinn2023reflexion} built on ReAct by adding an LLM critic that reflects on failures and provides persistent verbal feedback to the agent for improved performance in subsequent episodes. The performance gains from these LLM agent methods are solely dependent on enhanced prompt-based in-context learning, which limits their ability for longer-term true learning from new varied experiences. While MUSE also makes use of prompting for both reasoning and planning, it can also continually learn from its experiences to facilitate more effective problem-solving in unknown situations and environments.

 \section{Decoder-based World Model implementation}
In this section, we describe our implementation of the MUSE framework using a decoder-based World Model to equip MBRL agents with metacognitive abilities of self-assessment and self-regulation. 

\subsection{Methods}
\subsubsection{Self-Assessment through World Modeling}
We leverage the decoder-based World Model from Dreamer-v3~\citep{hafner2023mastering} to implement self-assessment for agents, but we note that our approach can be extended to decoder-free World Models as well. Dreamer-v3 uses a Recurrent State-Space Model (RSSM) to model the environment dynamics. See Equations~\ref{eq:rssm}-\ref{eq:wm_recon} for the formulation from~\citet{hafner2023mastering}. The RSSM maps the input state $x_t$ and recurrent state $h_t$ to a latent embedding $z_t$ and uses the concatenation of $h_t$ and $z_t$, called the RSSM state, as input to parameterize various distributions over the reward $\hat{r}_t$, terminal signal $\hat{d}_t$, and decoded state $\hat{x}_t$. 

\eq{
\begin{alignedat}{4}
\raisebox{1.95ex}{\llap{\blap{\ensuremath{
\text{RSSM} \hspace{1ex} \begin{cases} \hphantom{A} \\ \hphantom{A} \\ \hphantom{A} \end{cases} \hspace*{-2.4ex}
}}}}
& \text{Sequence model:}        \padspace && h_t            &\ =    &\ f_\phi(h_{t-1},z_{t-1},a_{t-1}) \\
& \text{Encoder:}   \padspace && z_t            &\ \sim &\ \qp(z_t | h_t,x_t) \\
& \text{Dynamics predictor:}   \padspace && \hat{z}_t      &\ \sim &\ \pp(\hat{z}_t | h_t) \\
& \text{Reward predictor:}       \padspace && \hat{r}_t      &\ \sim &\ \pp(\hat{r}_t | h_t,z_t) \\
& \text{Terminal signal predictor:}     \padspace && \hat{d}_t      &\ \sim &\ \pp(\hat{d}_t | h_t,z_t) \\
& \text{Decoder:}        \padspace && \hat{x}_t      &\ \sim &\ \pp(\hat{x}_t | h_t,z_t)
\end{alignedat}
\label{eq:rssm}}

Following~\citet{hafner2023mastering}, given a sequential batch of inputs $x_{1:T}$, actions $a_{1:T}$, rewards $r_{1:T}$, and terminal signals $d_{1:T}$, the World Model parameters $\phi$ are optimized to minimize the prediction loss $\mathcal{L}_{\mathrm{pred}}$, the dynamics loss $\mathcal{L}_{\mathrm{dyn}}$, and the representation loss $\mathcal{L}_{\mathrm{rep}}$. The prediction loss $\mathcal{L}_{\mathrm{pred}}$ is the joint negative log-likelihood of the multiple probabilistic predictors (Equation~\ref{eq:wm_recon}). Real-valued quantities like the reward and decoded state are trained with a symlog squared loss, whereas the terminal signal, which is a binary-valued quantity, is trained with logistic regression. The dynamics loss $\mathcal{L}_{\mathrm{dyn}}$ and the representation loss $\mathcal{L}_{\mathrm{rep}}$ are designed to effectively learn the dynamics of the latent embeddings for generating realistic rollout trajectories. 

\eq{
\mathcal{L}(\phi)\doteq
\E_{q_\phi}<\Bigg>[\displaystyle\sum_{t=1}^T(
    \mathcal{L}_{\mathrm{pred}}(\phi)
   +\mathcal{L}_{\mathrm{dyn}}(\phi)
   +0.1 \mathcal{L}_{\mathrm{rep}}(\phi)
)]
\label{eq:wm}
}
\eq{
\mathcal{L}_{pred} \doteq -\ln \pp(r_t | h_t, z_t) - \ln \pp(d_t | h_t, z_t) - \ln \pp(x_t | h_t, z_t)
\label{eq:wm_recon}
}

\begin{figure}[h!]
    \centering
    \includegraphics[width=0.6\linewidth]{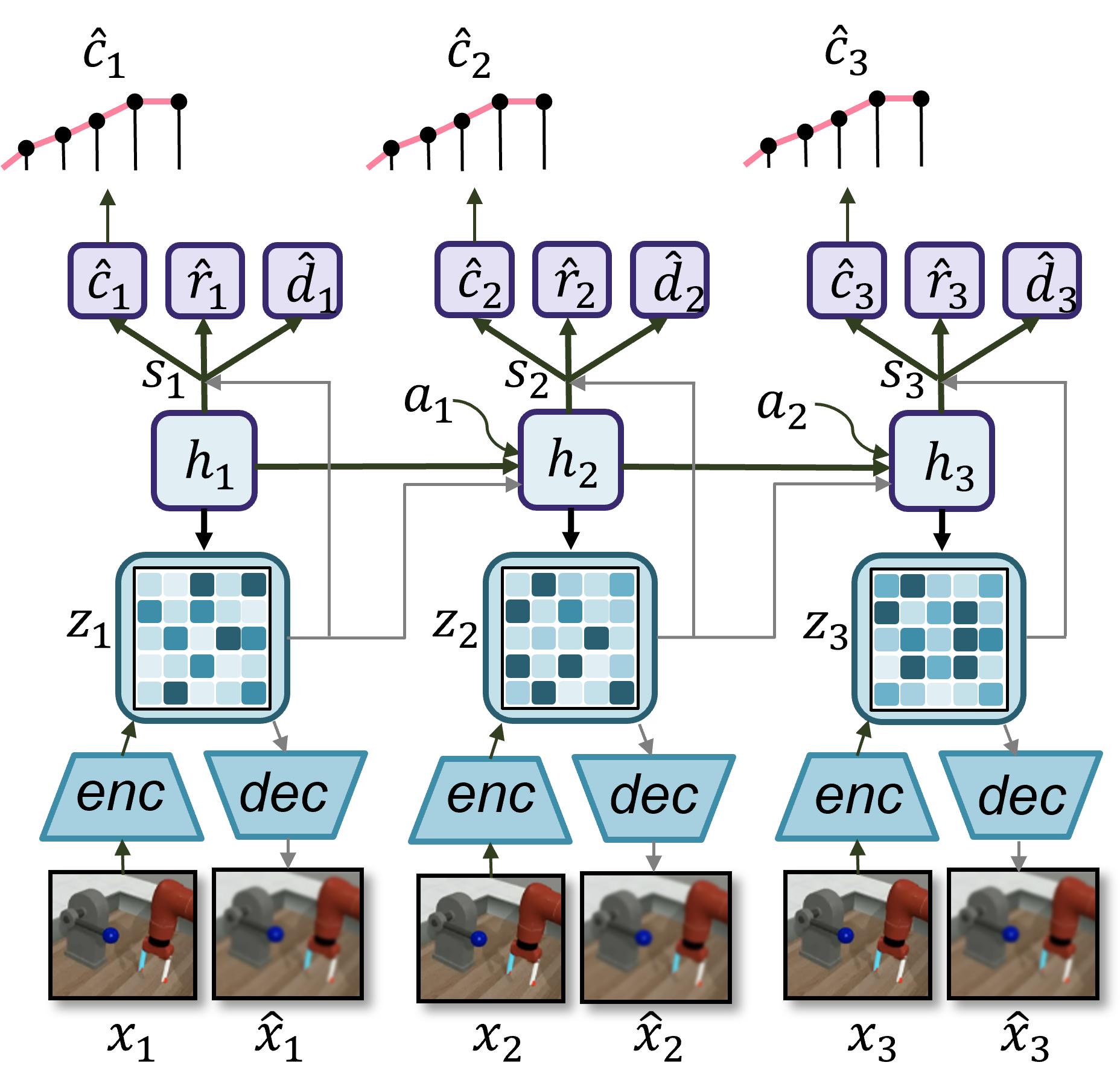}
    \caption{Schematic of the implementation of self-assessment in the context of the Dreamer-v3 World Model (adapted from~\citet{hafner2023mastering}). The input state $x$ to the RSSM is encoded into latent embedding $z$. The model recurrently predicts self-assessment $\hat{c}$, reward $\hat{r}$, and terminal signal $\hat{d}$, while also decoding the input state $\hat{x}$.}
    \label{fig:self-awareness-arch}
\end{figure}

We augment Dreamer-v3's World Model with an additional head for predicting task success. In this implementation, the Self-Assessment Model is an MLP with $N$ outputs that map the RSSM state to the probability of task success within the $N$ quantiles of the maximum episode duration. Specifically, the MLP outputs parameterize $N=5$ Bernoulli distributions $\{\psi_1, \cdots \psi_N\}$ for the five quantiles. A self-assessment prediction involves sampling from each of these distributions, $\hat{c}^{i}_{t} \sim \psi_i(h_t, z_t)$, and combining these samples into a prediction vector. For example, a prediction of success in the first quantile would yield the vector $[1, 1, 1, 1, 1]$, whereas a prediction of failure even by the last quantile would yield the vector $[0, 0, 0, 0, 0]$. This process is visualized in Figure~\ref{fig:self-awareness-arch}. Note that each component of the self-assessment head is trained separately using binary cross-entropy loss, and the individual losses are then added to the total prediction loss $\mathcal{L}_{\mathrm{pred}}$ (Equation~\ref{eq:m_sa_eqs}).

\eq{
\mathcal{L}_{\mathrm{SA}} \doteq -\sum_{i=1}^N \ln \psi_i(h_t, z_t), \padspace \mathcal{L}_{pred} \leftarrow  \mathcal{L}_{pred} + \mathcal{L}_{\mathrm{SA}}
\label{eq:m_sa_eqs}
}

\subsubsection{Self-Regulation} \label{subsec:sr_options}
Even with pre-deployment training on multiple tasks, including parametric variations, MBRL agents exhibit limited generalization to novel tasks that require either new, orchestrated combinations of those skills or entirely new skills ~\citep{ketz2022}. While Dreamer-v3 can handle novel parametric variations for a known task, it struggles to make progress when faced with an unknown reward function that differs semantically from those of the training tasks. Central to our MUSE framework is the self-regulation algorithm, which performs competence-aware actions to solve novel tasks. Specifically, the decision-making process selects actions that maximize the likelihood of task success. Self-assessed competence can be used to guide planning in three primary ways: 
\begin{enumerate}[label=(\alph*)]
    \item Simulate multiple future scenarios (rollout trajectories) based on the current state and potential actions, then greedily select the path that maximizes the self-assessment criterion
    \item Perform MCTS over actions using the self-assessment criterion in place of variations of the UCB score
    \item Optimize the RSSM state to directly maximize the self-assessment criterion to effectively self-regulate the policy 
\end{enumerate}

For this implementation of MUSE, we used the third option. This self-regulation method, which is detailed in Algorithm~\ref{alg:self-regulation}, leverages the differentiability of the World Model with the self-assessment head. MUSE performs a World Model rollout where the agent's actions are regulated to increase the likelihood of reaching a success state. Specifically, MUSE directly optimizes the RSSM state, $s$ ($\doteq \{h,z\}$), input to the agent for selecting actions that maximize the self-assessment criterion. The intuition is that when the Self-Assessment Model predicts failure in a novel environment, it is no longer useful to just rely on the default policy. Instead, we seek competence-aware actions that guide the agent to a success state by augmenting the RSSM state in a direction that increases the probability of task success. At each time step $t$, MUSE performs a short World Model rollout of horizon $H$ that optimizes over the RSSM state $s$ as follows: 
\eq{
s \gets s + \beta \nabla_{s}\left(\sum_{i=1}^N\psi_i(s)\right). 
}
It then applies the self-regulated action $a_t \sim \pi(a_{t} | s)$, observes the resulting recurrent state $h_{t+1}$, and begins a new iteration.

\begin{algorithm}
\caption{Self-regulation leverages the World Model and its constituent Self-Assessment Model to select competence-aware actions}
\begin{algorithmic}[1]
\Require $h_{t}$ 
\State $H$ = 10
\State $\beta = 0.02$ 
\State $z_t \sim p_\phi(z_t | h_t)$
\State $h \gets h_{t}$
\State $z \gets z_{t}$
\State $s \doteq \{h,z\}$
\For{$step \gets 1$ to $H$}
    \State $a \sim \pi(a | s)$
    \State $h \gets f_\phi(s,a)$
    \State $ \displaystyle s \gets s + \beta \nabla_{s}\left(\sum_{i=1}^N\psi_i(s)\right)$
\EndFor
\State $a_{t} \sim \pi(a_{t} | s)$
\State $h_{t+1} \gets f_\phi(s,a_{t})$
\State \textbf{return} $z_{t}$, $a_{t}$, $h_{t+1}$
\end{algorithmic}
\label{alg:self-regulation}
\end{algorithm}

\subsection{Meta-World Experiments}
For the World-Model-based implementation, we evaluated our approach within the Meta-World robotic manipulation simulator~\citep{yu2020meta} and compared it against Dreamer-v3~\citep{hafner2023mastering} as the MBRL baseline. Meta-World provides a suitable testbed for learning a shared perceptual and dynamics model across multiple tasks using a 6 degrees-of-freedom (DOF) robotic arm. 

To ensure consistency, we used Dreamer-v3’s network architectures, hyperparameters, and learning procedures across all shared components between the two agents (e.g., an imagination horizon of 15 time steps for actor and critic learning). Both methods were implemented in the same PyTorch codebase, with the self-assessment and self-regulation modules omitted for Dreamer-v3. The agents received a $64 \times 64$ RGB observation alongside a 40-dimensional proprioceptive state. Additionally, we included a task embedding that was represented as a single integer-valued channel appended to the visual state. MUSE leveraged the built-in \texttt{success} signal returned by the Meta-World environment to train its Self-Assessment Model. For the experiments, we employed a two-stage protocol comprising pre-deployment training on known tasks followed by deployment adaptation to unknown tasks. Each episode in these experiments had a maximum time limit of 500 steps.

\begin{figure}[h!]
    \centering
    \includegraphics[width=12cm]{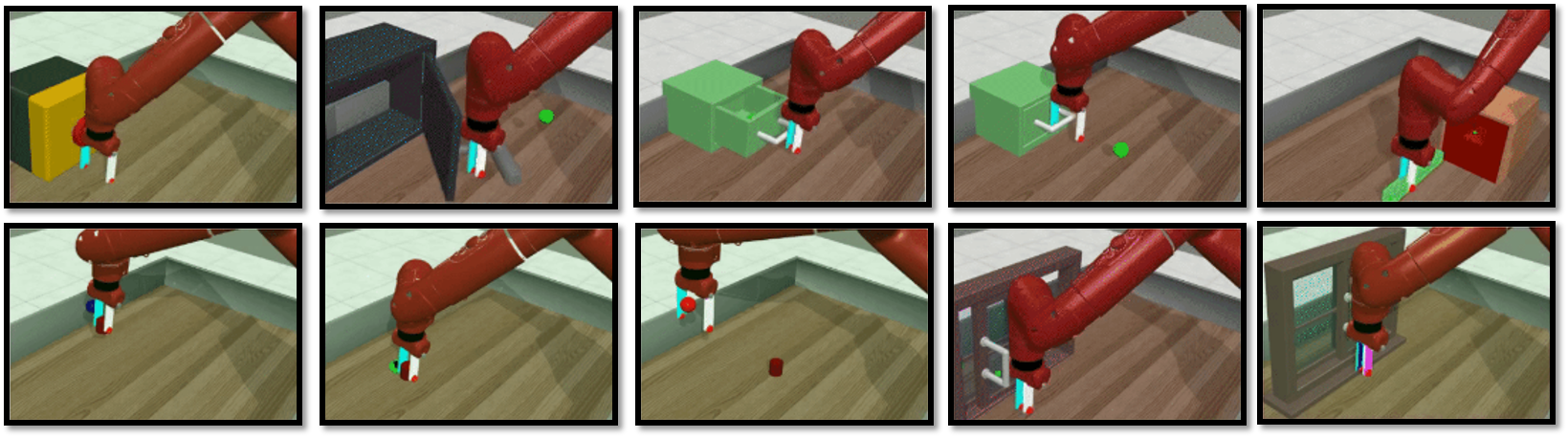}
    \caption{Meta-World pre-deployment training set \texttt{
[button-press, door-open, drawer-close, drawer-open, peg-insert-side, pick-place, push, reach, window-close, window-open]}, which comprises the 10 tasks from the MT10 suite.}
    \label{fig:metaworld-mt10}
\end{figure}

\begin{figure}[h!]
    \centering
    \includegraphics[width=12cm]{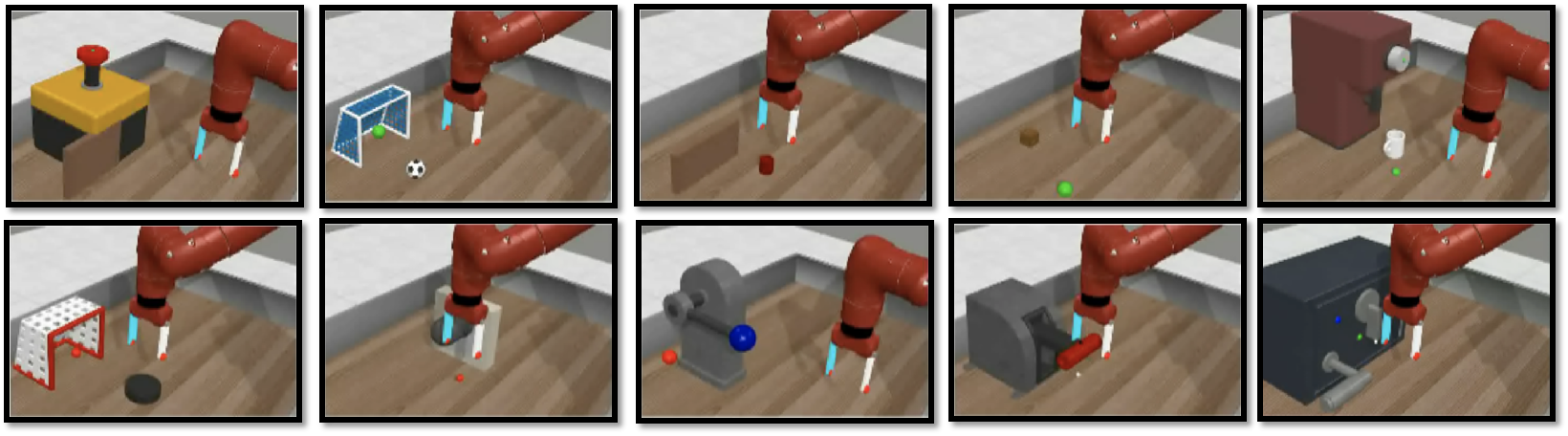}
    \caption{Meta-World evaluation set \texttt{
[button-press-topdown-wall, soccer, push-wall, push-block, coffee-button, plate-slide, peg-unplug-side, lever-pull, handle-press, door-unlock]}, which comprises 10 novel tasks from the MT50 suite with distinct reward functions, which differ semantically from those in the pre-deployment training set.}
    \label{fig:metaworld-eval}
\end{figure}

For pre-deployment training, we utilized Meta-World's MT10 suite of 10 different manipulation tasks (Figure~\ref{fig:metaworld-mt10}). In particular, we adopted a multi-task learning paradigm~\citep{mandi2023cacti}, encompassing all 10 training tasks over 2M total environment steps. This paradigm was chosen over meta-RL approaches to reduce computational costs and training time~\citep{wang2021}. By default, object and goal positions were randomly sampled to enable domain randomization. For deployment adaptation, we evaluated the agents on a set of 10 novel tasks from Meta-World's MT50 suite with distinct reward functions, which were semantically different from those in the pre-deployment training set (Figure~\ref{fig:metaworld-eval}). The agents were exposed to one novel task at a time, starting with pre-deployment trained weights, for 20 adaptation episodes per task. They were assessed for performance on novel tasks during these adaptation episodes. The task embedding channel was set to zero for novel tasks. During training as well as adaptation, both agents continually updated their World Models with real data and their actor and critic neural networks with imagined data. The replay buffer from pre-deployment training was retained for deployment adaptation to prevent catastrophic forgetting of previously trained performance.

\subsubsection{Metrics}
\underline{Self-Assessment}\\
We used metacognitive accuracy and the Area under the Type 2 Receiver Operating Characteristic Curve (AUROC2) to evaluate how well MUSE predicts its success on novel tasks~\citep{fleming2014}, which in turn indicates how effectively the self-assessment signal can support online adaptation. MUSE was evaluated for each novel task separately over 20 adaptation episodes. During these episodes, we collected MUSE's step-wise self-assessment predictions for evaluation. The predictions at all time steps across tasks were compared with the true labels to compute the metacognitive accuracy and AUROC2 metrics. The Type 2 ROC curve for MUSE was computed by treating the sum of the self-assessment components $\left(\sum_{i=1}^N\psi_i(s)\right)$ as the signal. Note that, as Dreamer-v3 lacks a self-assessment module to predict episode outcomes, we cannot make a meaningful comparison between Dreamer-v3 and MUSE for self-assessment. 

\underline{Self-Regulation}\\
We evaluated how well the agents generalize to unknown situations using two metrics: the percentage of novel tasks solved and the average time to task completion. Each agent was assessed for each novel task separately over the 20 adaptation episodes. The percentage of episodes where the agent completed the task within the maximum time limit was averaged across all novel tasks to calculate the success rate. Similarly, the number of time steps required to achieve success in each episode was averaged across all episodes for each novel task to compute the respective time-to-completion metrics. 

\begin{table}[h!]
\centering
\caption{Self-assessment performance of the MUSE agent on novel tasks in the Meta-World environment.}
\label{table:self_awareness_table}
\begin{tabular}{lcc}
\hline
Method & Metric & Value \\ 
\hline
MUSE  & Metacognitive Accuracy & 92\% \\ 
MUSE  & AUROC2 & 0.95 \\ 
\hline
\end{tabular}
\end{table}

\subsubsection{Results}
\begin{figure}[h!]
    \centering
    \includegraphics[width=\linewidth]{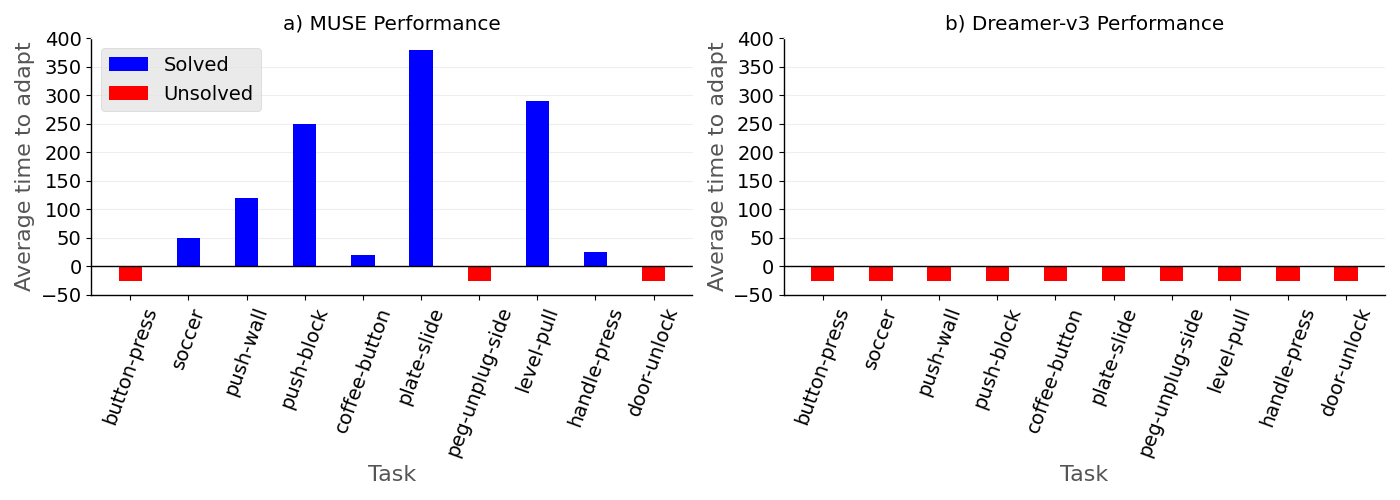}
    \caption{Average number of time steps per episode required to solve each of the 10 novel tasks in the Meta-World environment. MUSE (a) successfully solved 7 out of 10 tasks, whereas Dreamer-v3 (b) failed to solve any of them. Note that to facilitate illustration, unsolved tasks are assigned a nominal value of -25 and depicted by a red bar.}
    \label{fig:self-regulation_adaptation-steps}
\end{figure}

The MUSE agent achieved a metacognitive accuracy of 92\% and an AUROC2 of 0.95 on novel tasks over the adaptation episodes, demonstrating that the Self-Assessment Model is highly predictive of competence for novel tasks (Table~\ref{table:self_awareness_table}). Further, MUSE successfully solved 7 of the 10 novel tasks (70\%) by leveraging competence-aware actions. The 7 solved tasks required different time steps per episode depending on the relative difficulty and complexity of the task (Figure~\ref{fig:self-regulation_adaptation-steps}a). For instance, \texttt{plate-slide} required nearly the maximum time budget, while \texttt{coffee-button} was solved quickly. But in sharp contrast, Dreamer-v3 failed to solve any of the novel tasks within the allotted adaptation episodes (Figure~\ref{fig:self-regulation_adaptation-steps}b). 

\subsection{Discussion}
The pre-deployment training tasks were selected to encompass a broad range of skills, so the agents could develop a better understanding of various manipulation strategies. Following this training, the agents were exposed to out-of-distribution tasks that were selected to evaluate their ability to generalize learned skills to novel challenges. This two-stage approach of training on familiar tasks and then adapting to novel ones was used to rigorously assess the effectiveness of MUSE and Dreamer-v3 in unknown situations. Overall, the World-Model-based experiments showed that MUSE significantly outperforms Dreamer-v3 in adapting to novel tasks by effectively leveraging competence awareness for strategy selection.

\section{LLM-based implementation}
In this section, we describe our implementation of the MUSE framework to equip LLM agents with metacognitive abilities of self-assessment and self-regulation.

\subsection{Methods}
\subsubsection{ReAct}
\begin{figure}[h!]
    \centering
    \includegraphics[keepaspectratio,width=0.8\textwidth]{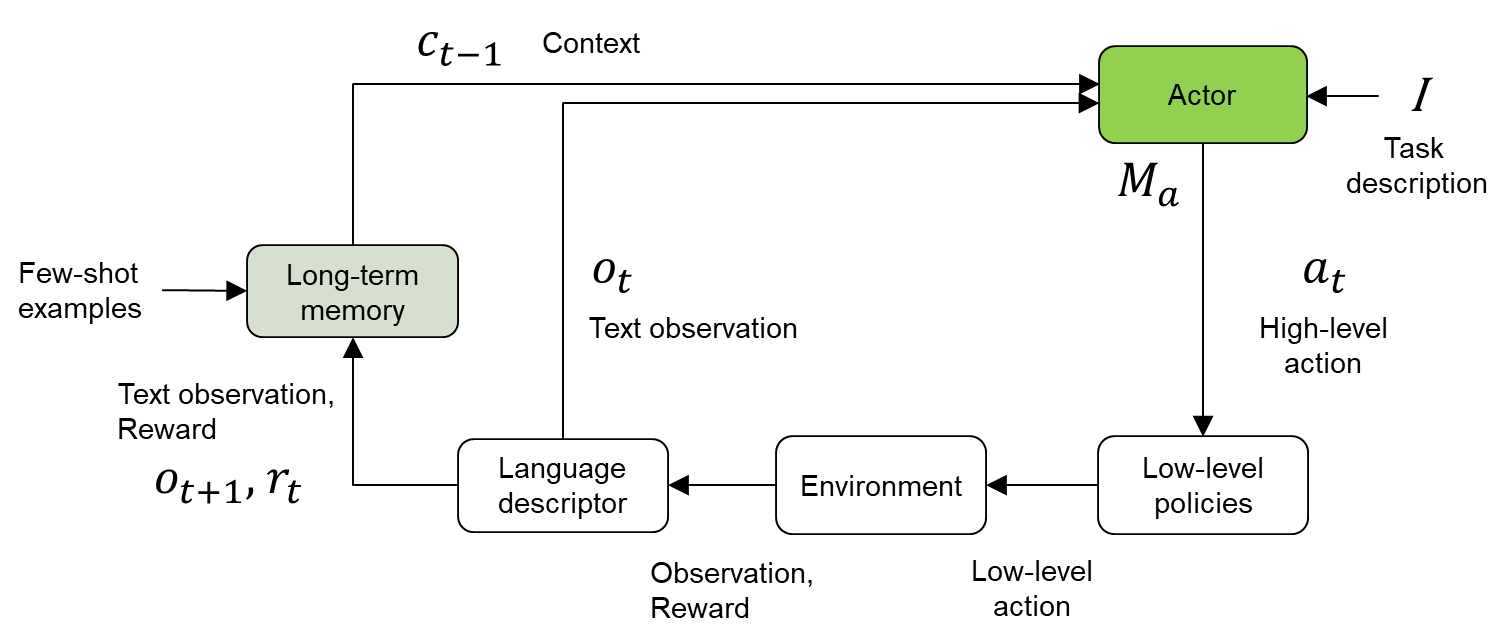}
    \caption{Our illustration of the ReAct architecture~\citep{yao2023react}.}
    \label{fig:ReAct}
\end{figure}

~\citet{yao2023react} was among the first to introduce an LLM agent that interacts with its environment to accomplish tasks by being prompted to reason and act (Figure~\ref{fig:ReAct}). At time step \(t\), the ReAct agent ($M_a$) perceives an observation \(o_t \in \mathcal{O}\), executes an action \(a_t \in \mathcal{A}\) based on the policy \(\pi(a_t | I, c_{t-1}, o_t)\), and receives a reward $r_t$. Here, $I$ is the natural language description of the task, and \(c_t = \{o_1, a_1, r_1 ..., o_{t}, a_{t}, r_{t}\}\) is the running context of the trajectory within the episode. The context is reset at the end of each episode, which occurs when either the task is solved within the time budget or the episode terminates due to running out of time. The ReAct agent leverages language-based CoT reasoning ~\citep{wei2022chain} for improved sequential action selection for multi-step tasks. Note that actions $a_t$ stored in $c_t$ include both standard actions and CoT reasoning steps. Two-shot domain-specific trajectory examples with step-by-step reasoning are included in the prompt for CoT reasoning. See Supplementary Section~\ref{appendix:prompting} for illustrative examples. 

\subsubsection{Reflexion}
\begin{figure}[h!]
   \centering
    \includegraphics[keepaspectratio,width=0.8\textwidth]{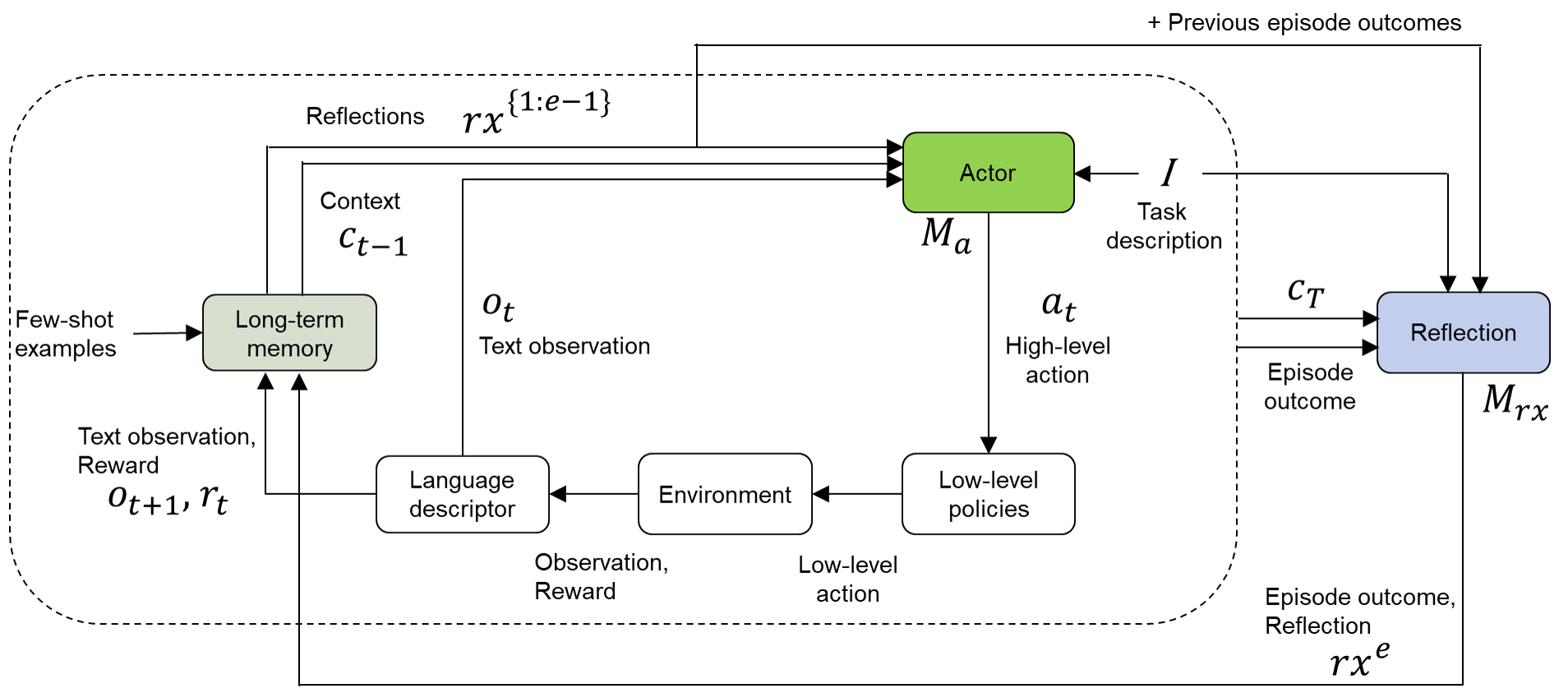}
    \caption{Our illustration of the Reflexion architecture~\citep{shinn2023reflexion}, which leverages verbal RL to enable agents to learn from past episodes.}
    \label{fig:Reflex}
\end{figure}

Reflexion~\citep{shinn2023reflexion} enhances in-context prompting for LLM agents by generating internal feedback in language, referred to as ``reflection,'' to transfer lessons learned across episodes for a given task (Figure~\ref{fig:Reflex}). At the end of each episode \(e\), Reflexion prompts an LLM to reason and generate verbal feedback \(rx^e\) about the agent's performance by analyzing the entire trajectory \(c_T\) from the episode start ($t=1$) to finish ($t=T$) and the episode outcome (success or failure). This Reflection LLM ($M_{rx}$) also receives the reflections and outcomes from previous episodes ($1:e-1$). The Reflexion agent ($M_a$) follows the policy \(\pi(a_t | I, c_{t-1}, o_t, rx^{\{1:e-1\}} )\), which incorporates cumulative reflections from earlier episodes. In contrast to ReAct, which relies solely on short-term memory comprising the agent's running trajectory within the current episode, Reflexion leverages both short- and long-term memory for improved strategic adaptation to novel situations. This iterative internal feedback mechanism allows the agent to refine its strategies progressively by integrating lessons learned from failures in previous episodes. Reflexion addresses the credit assignment problem through implicit reasoning about specific actions within the trajectory that led to failures, proposing alternative strategies for future episodes. Through this process, the agent develops an enhanced understanding of effective plans to iteratively improve its adaptability to novel tasks. 

While the original Reflexion study~\citep{shinn2023reflexion} utilized a single LLM for both action generation and reflection, our implementation uses distinct LLMs that are fine-tuned for each respective task. Additionally, we extend the reflection process to include successful outcomes, which enables the model to reinforce effective strategies alongside learning from failures. 

\subsubsection{MUSE}
\begin{figure}[h!]
   \centering
    \includegraphics[keepaspectratio,width=\textwidth]{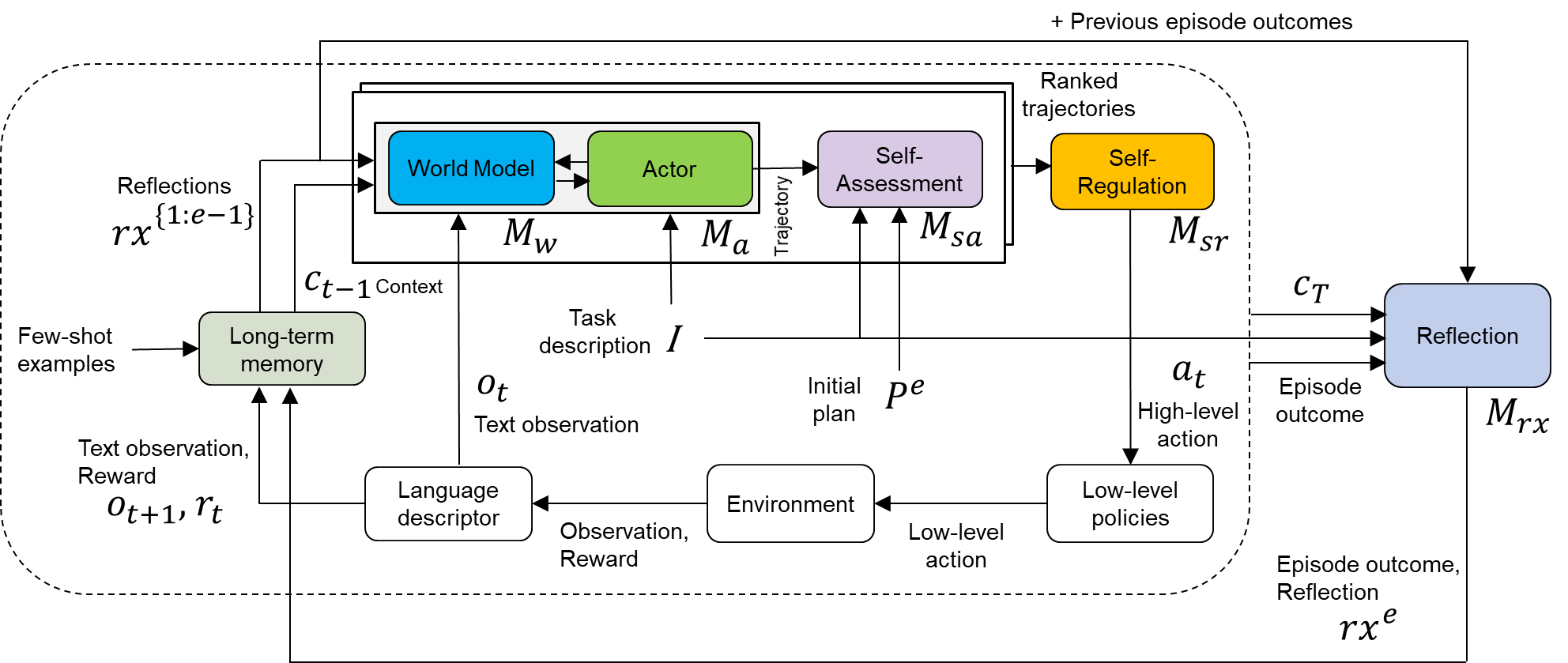}
    \caption{Illustration of the MUSE architecture for LLM agents, which implements the metacognitive cycle to iteratively solve unknown tasks.}
    \label{fig:MUSE}
\end{figure}

The MUSE framework, illustrated in Figure~\ref{fig:MUSE}, builds on mechanisms from ReAct and Reflexion by incorporating additional modules for self-assessment and self-regulation:
\begin{itemize}
    \item \textbf{World Model} (\(M_w\)): An LLM that can be prompted to predict the next observation, reward, and terminal signal given the current observation and action. This LLM works in conjunction with the Actor ($M_a$) to generate potential future states and actions (rollout trajectories), enabling look-ahead planning. In text-based domains, such as the experiments presented here, \(M_w\) can use the same LLM as \(M_a\). In fact, the same LLM can be directly prompted to generate diverse trajectories without requiring explicit interaction between the two.
    \item \textbf{Self-Assessment Model} (\(M_{sa}\)): A language-conditioned neural network that evaluates and scores trajectories generated by the World Model/Actor to assess their alignment and effectiveness for the agent's goals. Specifically, it predicts task competence, or the probability of task success, for each trajectory. 
    \item \textbf{Self-Regulation} (\(M_{sr}\)): A module that decides a competence-aware course of action based on one of the options outlined in Subsection~\ref{subsec:sr_options}. For this implementation of MUSE, \(M_{sr}\) chooses the first action $a_t$ from the rollout trajectory that is most likely to achieve task success, as determined by competence evaluations from \(M_{sa}\). 
\end{itemize}

\underline{World Model/Actor}\\ 
During deployment, the World Model/Actor generates several potential future state-action sequences with a horizon $H$ at each time step $t$. The diversity of these rollout trajectories, denoted by \(\tau_t = \{a_t, r_t, o_{t+1}, \dots, a_{t+H-1}, r_{t+H-1}, o_{t+H}\}\), is controlled by the temperature setting of the LLM. These trajectories represent hypothetical paths extending from the current observation $o_{t}$ and context $c_{t-1}$, guided by the task description $I$ and task-specific reflections $rx^{\{1:e-1\}}$ stored in memory. For this implementation, we explicitly did not specify a horizon $H$; instead, the LLM was allowed to generate trajectories up to the maximum length permitted by its context window. The temperature of the LLM was set to 0.5, and five rollout trajectories were generated at each time step. 

\underline{Self-Assessment}\\ 
The Self-Assessment Model (\(M_{sa}\)) utilizes a transformer encoder $\mathcal{M}$~\citep{SentenceTransformers} and an MLP $g_\eta$ to predict the probability of task success (Equation~\ref{eq:m_sa}) for rollout trajectories generated by the World Model/Actor before their actual execution in the environment. Specifically, \(M_{sa}\) evaluates the alignment and effectiveness of potential trajectories for the task at hand. The algorithm for training \(M_{sa}\) and using it for evaluation is detailed in Algorithm~\ref{alg:evaluator}.

\eq{
y_{\text{pred}} = M_{sa}(I, P^e, \tau)
\label{eq:m_sa}
}
Here, \(\tau\) represents a trajectory and \(P^e\) denotes the initial language-based plan generated by the World Model/Actor LLM at the start of episode $e$ based on task description $I$. See Supplementary Material for several illustrative examples of the initial plan \(P^e\). The output layer of \(M_{sa}\) employs a sigmoid activation function to yield a probability of task success. A threshold of 0.5 is used to convert these probabilities into binary outcomes (success or failure). During pre-deployment training and deployment adaptation, \(M_{sa}\) is trained to minimize the binary cross-entropy loss between the predicted task success (\(y_{\text{pred}}\)) and the actual binary outcome (\(y\)) as follows:
\eq{
\mathcal{L}(y, y_{\text{pred}}) \doteq -\left[y \log(y_{\text{pred}}) + (1 - y) \log(1 - y_{\text{pred}})\right]
\label{eq:loss_m_sa}
}
To increase the number of training samples and mitigate overfitting, the training trajectories are segmented into non-overlapping chunks, each containing four action-observation pairs. For example, the initial chunk within an episode includes pairs ($a_1$, $o_1$) through ($a_4$, $o_4$), while the next spans ($a_5$, $o_5$) through ($a_8$, $o_8$), and so on. 

\begin{algorithm}
\caption{Training and evaluation of Self-Assessment Model $M_{sa}$}
\label{alg:evaluator}
\begin{algorithmic}[1]
\State \textbf{Input:} Dataset $\mathcal{D} = \{(I, P^e, \tau, y)\}$, transformer encoder $\mathcal{M}$, MLP $g_\eta$
\State \textbf{Output:} Trained evaluator $g_\eta$

\Function{TrainEvaluator}{$\mathcal{D}$, $\mathcal{M}$, $g_\eta$}
    \For{each $(I, P^e, \tau, y) \in \mathcal{D}$}
        \State $Z \gets \mathcal{M}(I)$ \Comment{Generate task embedding}
        \State $S \gets \mathcal{M}(\tau, P^e)$ \Comment{Generate trajectory and plan embedding}
        \State $X \gets \text{concat}(S, Z)$ 
        \State $y_{\text{pred}} \gets g_\eta(X)$ \Comment{Predict task success probability}
        \State Update $g_\eta$ parameters to minimize $\mathcal{L}(y, y_{\text{pred}})$ 
    \EndFor
    \State \textbf{return} Trained MLP $g_\eta$
\EndFunction
\vspace{1em}
\Function{EvaluateTrajectory}{$I$, $P^e$, $\tau$, $\mathcal{M}$, $g_\eta$}
    \State $Z \gets \mathcal{M}(I)$ 
    \State $S \gets \mathcal{M}(\tau, P^e)$
    \State $X \gets \text{concat}(S, Z)$ 
    \State $y_{\text{pred}} \gets g_\eta(X)$ 
    \State \textbf{return} $y_{\text{pred}}$
\EndFunction

\end{algorithmic}
\end{algorithm}

\underline{MUSE Framework}\\
The pre-deployment training and deployment adaptation procedures for the MUSE agent are detailed in Algorithms~\ref{alg:muse} and~\ref{alg:muse_deploy}, respectively. For pre-deployment training, we first train the Reflexion agent~\citep{shinn2023reflexion} on each of the training tasks separately to collect data. The constituent models of the MUSE agent are then trained with multi-task supervised learning to benefit from the diversity of experiences. Specifically, the Reflection LLM (\(M_{rx}\)) in MUSE is trained using Direct Preference Optimization (DPO) with a preference dataset~\citep{rafailov2024} that is created by comparing reflections of success and failure across the training tasks. A success (positive) reflection occurs when the agent fails in episode \( e_i \) but succeeds in episode \( e_{i+1} \) following the reflection. Conversely, a failure (negative) reflection happens when the agent succeeds in episode \( e_i \) but fails in episode \( e_{i+1} \) despite the reflection. DPO directly optimizes \(M_{rx}\) to generate generalizable reflections that maximize the likelihood of satisfying these preferences (i.e., increase the likelihood of success and decrease the likelihood of failure). This enables MUSE to iteratively improve its ability to recover from failures and adapt to novel tasks through a more effective reflection process. 

The Actor LLM (\(M_{a}\)) is trained using Supervised Fine-Tuning (SFT) with only successful episodes, for simplicity, across the training tasks. By imitating the behavior demonstrated in these episodes, \(M_{a}\) learns to align its policy with actions seen during successful episodes. This joint multi-task fine-tuning process equips MUSE with generalizable success-driven strategies. Note that we use Low-Rank Adaptation (LoRA) ~\citep{hu2021lora} for parameter-efficient fine-tuning of both the $M_a$ and $M_{rx}$ LLMs. Additionally, the Self-Assessment Model (\(M_{sa}\)) undergoes supervised learning with pertinent multi-task data, which maps trajectory chunks to corresponding episode outcomes. 

During deployment, when the agents encounter a novel task, they engage in sequential episodes until success is achieved. MUSE employs competence-aware planning, as described above, to choose its actions. Furthermore, MUSE can perform online updates to each of its constituent models as new data becomes available from the novel task. Note that the baseline agents, ReAct and Reflexion, can also respond adaptively to new experiences but rely only on in-context prompting and/or reflection. However, these methods do not involve updating model parameters, which leads to knowledge being stored only for the short term. This constraint hinders the transfer of knowledge across episodes and tasks. MUSE overcomes these limitations by continually updating the weights of its models, enabling more effective knowledge transfer.

\begin{algorithm}
 \caption{Procedure for pre-deployment training of the MUSE agent}
Training dataset ${D_{in}} \sim$ in-distribution tasks \\
Specify max. training episodes, max. steps
\begin{algorithmic}[1]
\Function{PreTrainModels}{$D_{in}$, $M_a$, $M_{rx}$, $M_{sa}$, $M_{w}$} 
    \State Initialize $M_{sa}$, $M_w$, memory buffer
    \For{each task $\sim {D_{in}}$}
        \State Initialize $M_a$, $M_{rx}$ for the Reflexion agent
        \State Add $I$ to memory buffer
        \State Set $t=0$, $e=0$
        \While{$e <$ max. training episodes}
             \While{$t <$ max. steps \& task-not-solved}
                \State Generate $a_t$ using $M_a$ and submit to simulator
                \State Obtain $o_{t+1}$ and $r_t$, append $o_t$, $a_t$, $r_t$ to $c_t$
                \State Increment $t$
            \EndWhile
            \State Generate $rx^e$ using $M_{rx}$
            \State Append episode outcome, $c_t$, $rx^e$ to memory buffer
            \State Increment $e$
        \EndWhile
    \EndFor
    \State Use all data in memory buffer to train $M_a$, $M_{rx}$, $M_{sa}$, $M_w$ for MUSE  
    \State \textbf{return} Trained $M_a$, $M_{rx}$, $M_{sa}$, $M_{w}$    
\EndFunction
\end{algorithmic}
\label{alg:muse}
\end{algorithm}

\begin{algorithm}
 \caption{Procedure for deployment adaptation of the MUSE agent}
Test dataset ${D_{out}} \sim$ out-of-distribution tasks \\
Specify max. adaptation episodes
\begin{algorithmic}[1]
\Function{AdaptModels}{$D_{out}$, $M_a$, $M_{rx}$, $M_{sa}$, $M_{w}$}  
    \State Initialize memory buffer
    \For{each task $\sim {D_{out}}$}
        \State Set $M_a$, $M_{rx}$, $M_{sa}$, $M_{w}$ to pre-deployment trained weights
        \State Add $I$ to memory buffer
        \State Set $t=0$, $e=0$
        \While{$e <$ max. adaptation episodes}
             \While{$t <$ max. steps \& task-not-solved}
                \State Generate rollout trajectories using $M_w$ and $M_a$
                \State Evaluate each trajectory using $M_{sa}$
                \State Select the best action $a_t$ and submit to simulator
                \State Obtain $o_{t+1}$ and $r_t$, append $o_t$, $a_t$, $r_t$ to $c_t$
                \State Increment $t$
            \EndWhile
            \State Generate $rx^e$ using $M_{rx}$
            \State Append episode outcome, $c_t$, $rx^e$ to memory buffer
            \State Increment $e$
        \EndWhile
        \State Use data in memory buffer to update $M_a$, $M_{rx}$, $M_{sa}$, $M_w$
        \State Test updated MUSE agent on the current task over the test episodes 
    \EndFor
\EndFunction
  \end{algorithmic}
  \label{alg:muse_deploy}
\end{algorithm}

\subsection{ALFWorld Experiments}
For the LLM-based implementation, we evaluated our approach within the ALFWorld simulator~\citep{shridhar2020alfworld} and compared it against ReAct~\citep{yao2023react} and Reflexion~\citep{shinn2023reflexion} as the baselines. ALFWorld is a synthetic, text-based game simulator with diverse interactive environments that challenge agents to solve multi-step tasks (e.g., ``Find two plates and put them in a cabinet''). A unique aspect of ALFWorld is its emphasis on spatial and commonsense reasoning, requiring agents to infer probable locations of common household items; for example, recognizing that plates are typically found in kitchens. This characteristic makes ALFWorld particularly well-suited for leveraging the pre-trained commonsense knowledge of LLMs.

For these experiments, we selected the Mistral-7B-Instruct-v0.2 as the primary LLM~\citep{jiang2023mistral}. This LLM features a substantial context window size of 32,000 tokens, allowing it to process large chunks of data in a single pass for improved in-context learning. Similar to the World-Model-based experiments, we employed a two-stage protocol comprising pre-deployment training on known tasks followed by deployment adaptation to unknown tasks. Each episode in these experiments had a maximum time limit of 50 steps. To further explore the robustness of our framework, we conducted additional experiments using smaller or less capable LLMs such as Mistral-7B-OpenOrca~\citep{mukherjee2023orca,OpenOrcaMistral7B} and OpenELM-3B-Instruct~\citep{mehta2024openelm}.

The MUSE agent underwent pre-deployment training using 140 household tasks from ALFWorld's in-distribution (seen) set. For deployment adaptation, we evaluated all agents on 134 household tasks from ALFWorld's out-of-distribution (unseen) set. The agents were exposed to one novel task at a time, starting with the pre-deployment trained weights (for MUSE) or pre-trained LLM weights (for ReAct and Reflexion), under two conditions: (1) five test episodes for each task without any adaptation and (2) five adaptation episodes followed by five test episodes for each task. The agents were assessed for performance on the novel tasks over the test episodes. Note that there were no weight updates to the MUSE agent during these episodes. While it is possible to continually update all constituent models of MUSE (namely, $M_w$, $M_a$, $M_{rx}$, $M_{sa}$) using new data acquired from adaptation episodes, we chose to update only the MLP portion ($g_\eta$) of $M_{sa}$ to minimize computational time without loss of any generality. We used a replay buffer that included both training data and new adaptation data to update \(M_{sa}\) without catastrophic forgetting of pre-deployment trained performance. 

\subsubsection{Metrics}
\underline{Self-Assessment}\\
We used metacognitive accuracy and AUROC2 to evaluate how well MUSE predicts its success on the novel tasks. MUSE was evaluated for each novel task separately over five test episodes, at the end of pre-deployment training and after five adaptation episodes. During each test episode, the trajectory was segmented into non-overlapping chunks, each containing four action-observation pairs. MUSE's chunk-wise self-assessment predictions were then collected for evaluation, and the predictions for all chunks across tasks were compared against the true labels to compute the metacognitive accuracy and AUROC2 metrics. Note that, as the baseline agents, ReAct and Reflexion, lack a self-assessment module to predict episode outcomes, we cannot make a meaningful comparison between them and MUSE for self-assessment. 

\underline{Self-Regulation}\\
We evaluated how well the agents generalize to unknown situations using two metrics: the percentage of the novel tasks solved and the average time to task completion. Each agent was assessed for each novel task separately over the five test episodes. The percentage of episodes in which the agent completed the task within the maximum time limit was averaged across all the novel tasks to calculate the success rate. Similarly, the number of time steps required to achieve success in each episode was averaged across all episodes and tasks to compute the time-to-completion metric. There was a penalty of 100 time steps for failures.

\subsubsection{Results}
\begin{figure}[h!]
\centering
\includegraphics[width=0.8\textwidth]{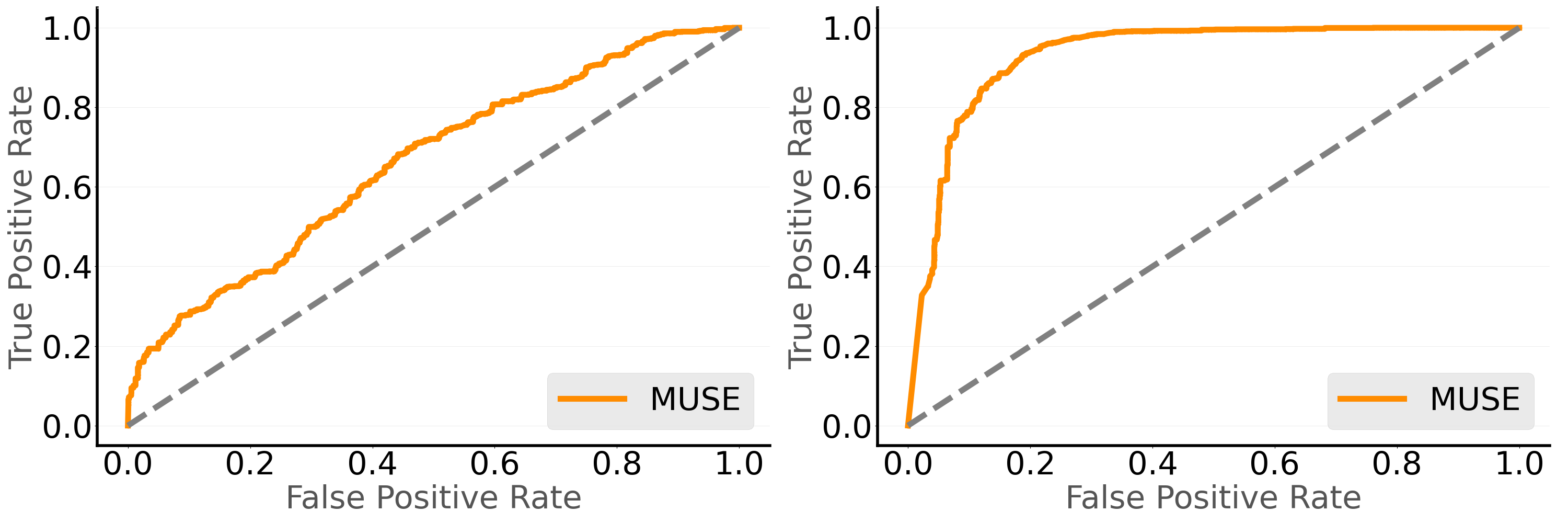}
\caption{Type 2 ROC curves of the MUSE agent for self-assessment of competence on the novel tasks in the ALFWorld environment, at the end of pre-deployment training (left) and after five adaptation episodes (right).}
\label{fig:self_awareness_roc_curve_2024}
\end{figure}

\underline{Self-Assessment}\\
The self-assessment results are summarized in Table~\ref{table:self_awareness_table_llm}, with the corresponding Type 2 ROC curves shown in Figure~\ref{fig:self_awareness_roc_curve_2024}. Following the five adaptation episodes, the MUSE agent achieved an AUROC2 of 0.93 and a metacognitive accuracy of 85\%, demonstrating a high level of competence awareness. 

\begin{table}[h!]
\centering
\caption{Self-assessment performance of the MUSE agent on the novel tasks in the ALFWorld environment, at the end of pre-deployment training (No Adaptation) and after five adaptation episodes (Adaptation).}
\label{table:self_awareness_table_llm}
\begin{tabular}{lcc}
\hline
Condition & AUROC2 & Metacognitive Accuracy \\
\hline
No Adaptation & 0.66 & 60\% \\
Adaptation      & 0.93 & 85\% \\  
\hline
\end{tabular}
\end{table}

\begin{table}[h!]
\centering
\caption{Self-Regulation: Success rate of various agents on the novel tasks, at the end of pre-deployment training (No Adaptation) and after five adaptation episodes (Adaptation).}
\label{table:self_regulation_table_llm}
\begin{tabular}{lcc}
\hline
Agent & No Adaptation & Adaptation \\
\hline
ReAct      & 35\% & 35\% \\
Reflexion  & 45\% & 51\% \\
Reflexion with supervised learning & 62\% & 65\% \\
MUSE with the lowest self-assessment & 59\% & 61\% \\
MUSE with random self-assessment & 68\% & 69\% \\
\textbf{MUSE}       & \textbf{84\%} & \textbf{90\%} \\
\hline
\end{tabular}
\end{table}

\begin{table}[h!]
\centering
\caption{Self-Regulation: Average number of time steps to success per episode for various agents on the novel tasks, at the end of pre-deployment training (No Adaptation) and after five adaptation episodes (Adaptation).}
\label{table:self_regulation_table_llm_steps}
\begin{tabular}{lcc}
\hline
Agent & No Adaptation & Adaptation \\
\hline
ReAct      & 97 & 97 \\
Reflexion  & 77 & 66 \\
Reflexion with supervised learning & 61 & 54 \\
MUSE with the lowest self-assessment & 65 & 62 \\
MUSE with random self-assessment & 53 & 49 \\
\textbf{MUSE}       & \textbf{43} & \textbf{38} \\
\hline
\end{tabular}
\end{table}

\begin{table}[h!]
\centering
\caption{Self-Regulation: Success rate of various agents with smaller or less capable LLMs on the novel tasks, after five adaptation episodes.}
\label{table:self_regulation_table_llm_other}
\begin{tabular}{lcc}
\hline
Agent & Mistral-7B-OpenOrca & OpenELM-3B-Instruct \\
\hline
ReAct      & 23\% & 4\% \\
Reflexion  & 27\% & 9\% \\
\textbf{MUSE}       & \textbf{58\%} & \textbf{55\%} \\
\hline
\end{tabular}
\end{table}

To highlight the importance of self-assessment, we provide examples of trajectories generated by different agents. As shown in Supplementary Section~\ref{appendix:muse_sampling}, the key to MUSE's success lies in its continual competence awareness. MUSE generates several rollout trajectories, assigns competence scores to each, and selects the one most likely to result in task success. For the given task, MUSE begins with an initial plan similar to that of Reflexion: ``find a bowl, clean it at the sink, and then place it in the cabinet.'' However, MUSE sets itself apart by engaging in deliberate actions that ensure steady progress toward the goal with increasing competence. After picking up the bowl from cabinet 1, MUSE evaluates multiple options, such as follows:
\begin{enumerate}[label={[\arabic*]}, start=0, noitemsep]
    \item ``think: Now I find a bowl (1). Next, I need to take it..OK..> take bowl 1 from cabinet 1.'' … Score: 0.6284
    \item ``think: Now I take a bowl (1). Next, I need to go to sinkbasin (1) and clean it.'' … Score: 0.8689
    \item ``think: Now I find a bowl (1). Next, I need to take it..OK..> take bowl 1 from cabinet 1. Nothing happens.'' … Score: 0.6417
\end{enumerate}

In this instance, MUSE selects Trajectory 1 due to its higher competence score, which indicates a greater likelihood of task success. By continually assessing its competence, MUSE adapts its strategy in real time to effectively navigate the environment's complexities. It maintains a coherent plan and avoids redundant or detrimental actions that could lead to confusion or failure. In contrast, Reflexion fails due to its lack of competence awareness. After acquiring the bowl, Reflexion redundantly decides to find and take a bowl again. This leads to a faulty sequence of actions, causing Reflexion to enter an incorrect state from which it cannot recover. Without competence awareness, Reflexion fails to recognize that it already possesses the bowl. Its actions become increasingly erratic, failing to contribute toward task completion. For the complete trajectory, refer to Supplementary Section~\ref{appendix:muse_sampling}.

In summary, competence awareness allows MUSE to adapt more effectively to unfamiliar situations. By avoiding low-competence rollout trajectories, it steers clear of actions leading to failure, promotes exploration within the agent's capabilities, and prevents the agent from getting stuck in unrecoverable states. This approach results in smarter exploration and more effective online learning in response to novel scenarios.

\underline{Self-Regulation}\\
The self-regulation results for MUSE and the two baseline agents are summarized in Tables~\ref{table:self_regulation_table_llm} and~\ref{table:self_regulation_table_llm_steps}. These results demonstrate that MUSE solves more novel tasks and requires fewer time steps to completion per episode, both at the end of pre-deployment training and after five adaptation episodes. Notably, MUSE achieved a high success rate of 90\% on the 134 novel tasks, outperforming the baseline agents of ReAct and Reflexion by >150\% and >75\%, respectively. It's important to note that all agents in the pertinent comparisons have access to the same data, ensuring a fair evaluation. ReAct and Reflexion can also respond adaptively to new experiences but rely only on in-context prompting and/or reflection. So, the observed performance gains for MUSE are from its ability to leverage available data more effectively by performing competence-aware planning, not from access to any privileged information. To clarify and validate this point, we conducted additional experiments comparing:
\begin{itemize}
    \item Reflexion with supervised learning across the training tasks during pre-deployment (namely, DPO and SFT)
    \item MUSE with self-regulation based on the lowest self-assessment
    \item MUSE with self-regulation based on random self-assessment
\end{itemize}

The self-regulation results for these control conditions are also summarized in Tables~\ref{table:self_regulation_table_llm} and~\ref{table:self_regulation_table_llm_steps}. Reflexion with supervised learning during pre-deployment can consolidate knowledge across the training tasks, but exhibits only a limited improvement over its original implementation~\citep{shinn2023reflexion} in adaptation to the novel tasks. This suggests that even a generalized reflection mechanism is less effective in novel scenarios without self-assessment and self-regulation. Furthermore, variants of MUSE using poor (random or pessimistic) self-assessment signals underperform, validating the importance of self-assessment for effective self-regulation. Note that even when self-assessment is flawed, the performance of the MUSE agent does not collapse because the Actor LLM ($M_{a}$) has benefited from pre-deployment training to already generate reasonable plans for competence-aware selection.

To further evaluate the contribution of the self-assessment and self-regulation framework in MUSE beyond LLM-powered reasoning, we conducted additional experiments using smaller or less capable LLMs. Table~\ref{table:self_regulation_table_llm_other} presents the post-adaptation performance of MUSE compared to the baseline agents when employing smaller or less capable LLM variants. MUSE maintained relatively higher success rates (55-58\%) on the novel tasks, achieving >2x and >6x better performance than Reflexion when using Mistral-7B-OpenOrca and OpenELM-3B-Instruct, respectively. These results suggest that metacognition plays a particularly significant role when operating under constrained conditions, such as with smaller or less capable LLMs.

\subsection{Discussion}
Through these experiments, we evaluated the metacognitive capabilities of MUSE to showcase its superior performance over the baseline agents of ReAct and Reflexion. With reliable competence awareness on the novel tasks, the MUSE agent can make informed decisions by avoiding rollout trajectories likely to lead to failure and prioritizing those with a higher probability of task success for more effective online adaptation. MUSE's ability to learn efficiently from a limited number of adaptation episodes is particularly critical in unknown situations and environments, where task complexity and novelty can vary significantly. Furthermore, its strong performance even with less-capable LLMs suggests that incorporating metacognitive components into mainstream AI agents can enhance performance under resource-constrained conditions. Overall, the LLM-based experiments demonstrated that integrating self-assessment and self-regulation into LLM-based agents can significantly improve their ability to handle unknown scenarios.

\section{Conclusions}
In this article, we introduced the Metacognition for Unknown Situations and Environments (MUSE) framework to integrate self-assessment and self-regulation into AI agents to enhance their adaptability to unknown situations and environments. MUSE uses self-assessed competence to intelligently guide an iterative trial-and-error process to identify the right strategy. In contrast, current AI agents either blindly execute their routine or policy, risking catastrophic failures, or freeze as a result of runtime anomaly detection. These limitations arise because big data-driven solutions inherently struggle to adapt to novel scenarios with little or no training data. 

We presented two implementations of MUSE, one for MBRL agents and another for LLM-based agents, to instantiate the metacognitive cycle. In both cases, our experiments demonstrated that MUSE agents outperform baseline methods in handling novel tasks. These results show that MUSE facilitates problem-solving in unfamiliar environments by vetoing potentially catastrophic solutions, guiding the search toward strategies that leverage the agent's strengths, and learning online from the environment with each attempt. In high-stakes applications like autonomous driving, MUSE can facilitate safety by relying on its self-assessment as a form of competence calibration. Furthermore, the online learning during deployment enables MUSE to fix and overcome potential metacognitive misjudgments in novel scenarios. This avoids the infinite regress of needing to calibrate the preceding lower-order calibration. In summary, by enabling agents to be aware of their competence and to regulate their strategies accordingly, MUSE offers a promising pathway toward developing more resilient and versatile AI systems.

\section{Future Work}
While MUSE outperforms the baseline AI agents, metacognition in AI remains a nascent and exploratory field. The scope of metacognitive abilities is broad, offering several promising directions for future research. For instance, MUSE could be expanded to include metacognitive judgments that further enhance an agent's performance and learning. These include the Feeling of Knowing to help the agent persist in difficult, novel scenarios; the Judgment of Learning to facilitate resource-efficient learning; and Source Monitoring to assess the reliability of its knowledge. Another key avenue is integrating MUSE with lifelong learning techniques, such as generative experience replay and selective plasticity, to incrementally update its constituent models in new scenarios while mitigating catastrophic forgetting~\citep{rostami2019, kolouri2020, kudithipudi2023}. Such an integration would enable agents to retain and build upon previously learned knowledge while continually acquiring new skills, without requiring the storage of data from all prior conditions in the replay buffer, as was done in the current study.

Another exciting direction for future research involves applying MUSE to real-world domains, such as autonomous driving and robotics in unstructured settings. Investigating these domains for tasks with longer time horizons and larger state and action spaces will provide valuable insights into the scalability, robustness, and practicality of the MUSE framework. In this regard, MCTS methods can be adapted to leverage the self-assessment model from MUSE as the node evaluation function for guiding the search, enabling more effective exploration and problem-solving in complex environments. Another compelling avenue is expanding the metacognitive framework to incorporate additional aspects of human cognition that are currently absent in AI systems. For instance, integrating Theory-of-Mind could enhance agents' abilities to understand and predict human behaviors, enabling more effective collaboration in social or teamwork-driven environments. Additionally, exploring multi-agent interactions within the MUSE framework offers significant potential. Enabling multiple metacognitive AI agents to interact, collaborate, negotiate, or compete could lead to more sophisticated and efficient problem-solving at the group level. 
 
Our research highlights the potential of psychology- and brain-inspired approaches to address the limitations of current AI systems, particularly their dependence on extensive training data. By equipping AI agents with greater flexibility and adaptability, MUSE paves the way for more resilient and versatile systems capable of operating across a wide range of environments. We anticipate that future research will build upon and refine these ideas, advancing the field of metacognitive AI and expanding its applications across diverse domains.

\section{Acknowledgments}
The authors would like to thank Neale Ratzlaff for his invaluable contributions to this research and his constructive feedback.

\raggedbottom

\section{Supplementary Material}
Note that the trajectory examples provided below for the LLM-based experiments are exact copies of the LLM outputs. 

\subsection{Illustrative ALFWorld trajectories for in-context prompting.~\citep{yao2023react}}
\label{appendix:prompting}
Interact with a household to solve a task. Here are two examples. \\
// \\
{\small You are in the middle of a room. Looking quickly around you, you see a cabinet 13, a cabinet 12, a cabinet 11, a cabinet 10, a cabinet 9, a cabinet 8, a cabinet 7, a cabinet 6, a cabinet 5, a cabinet 4, a cabinet 3, a cabinet 2, a cabinet 1, a coffeemachine 1, a countertop 1, a diningtable 1, a drawer 1, a fridge 1, a garbagecan 1, a microwave 1, a shelf 3, a shelf 2, a shelf 1, a sinkbasin 1, a stoveburner 4, a stoveburner 3, a stoveburner 2, a stoveburner 1, and a toaster 1.}\\
{\small Your task is to: put a hot apple in fridge.}\\
{\small >think: To solve the task, I need to find and take an apple, then heat it with microwave, then put it in fridge.}\\
{\small OK.}\\
{\small >think: First I need to find an apple. An apple is more likely to appear in fridge (1), diningtable (1), coffeetable (1), drawer (1), cabinet (1-13), garbagecan (1). I can check one by one, starting with fridge 1.}\\
{\small OK.}\\
{\small >go to fridge 1}\\
{\small The fridge 1 is closed.}\\
{\small >open fridge 1}\\
{\small You open the fridge 1. The fridge 1 is open. In it, you see a cup 1, and a egg 1.}\\
{\small >go to diningtable 1}\\
{\small On the diningtable 1, you see a apple 1, a bread 1, a fork 3, a fork 2, a fork 1, a mug 2, a peppershaker 3, a plate 2, a pot 1, a soapbottle 3, a spatula 1, a spoon 3, a spoon 2, a spoon 1, a winebottle 3, and a winebottle 2.}\\
{\small >think: Now I find an apple (1). Next, I need to take it.}\\
{\small OK.}\\
{\small >take apple 1 from diningtable 1}\\
{\small You pick up the apple 1 from the diningtable 1.}\\
{\small >think: Now I take an apple (1). Next, I need to go to a microwave (1) and heat it.\\
OK.}\\
{\small >go to microwave 1}\\
{\small The microwave 1 is closed.}\\
{\small >heat apple 1 with microwave 1}\\
{\small You heat the apple 1 using the microwave 1.}\\
{\small >think: Now I heat an apple (1). Next, I need to put it in/on fridge 1.}\\
{\small OK.}\\
{\small >go to fridge 1}\\
{\small The fridge 1 is open. In it, you see a cup 1, and a egg 1.}\\
{\small >put apple 1 in/on fridge 1}\\
{\small You put the apple 1 in/on the fridge 1.}\\
// \\
{\small You are in the middle of a room. Looking quickly around you, you see a cabinet 10, a cabinet 9, a cabinet 8, a cabinet 7, a cabinet 6, a cabinet 5, a cabinet 4, a cabinet 3, a cabinet 2, a cabinet 1, a coffeemachine 1, a countertop 3, a countertop 2, a countertop 1, a diningtable 1, a drawer 6, a drawer 5, a drawer 4, a drawer 3, a drawer 2, a drawer 1, a fridge 1, a garbagecan 1, a microwave 1, a sinkbasin 1, a stoveburner 4, a stoveburner 3, a stoveburner 2, a stoveburner 1, and a toaster 1.}\\
{\small Your task is to: heat some egg and put it in diningtable.}\\
{\small >think: To solve the task, I need to find and take an egg, then heat it with microwave, then put it in diningtable.}\\
{\small OK.}\\
{\small >think: First I need to find an egg. An egg is more likely to appear in fridge (1), countertop (1-3), diningtable (1), stoveburner (1-4), toaster (1), garbagecan (1), cabinet (1-10). I can check one by one, starting with fridge 1.}\\
{\small OK.}\\
{\small >open fridge 1}\\
{\small You open the fridge 1. The fridge 1 is open. In it, you see a lettuce 2, a mug 2, and a potato 3.}\\
{\small >go to countertop 1}\\
{\small On the countertop 1, you see a bread 1, a fork 1, and a saltshaker 1.}\\
{\small >go to countertop 2}\\
{\small On the countertop 2, you see nothing.}\\
{\small >go to countertop 3}\\
{\small On the countertop 3, you see a bowl 1, a butterknife 1, a egg 2, a kettle 2, a plate 1, a sink 1, and a spatula 2.}\\
{\small >think: Now I find an egg (2). Next, I need to take it.}\\
{\small OK.}\\
{\small >take egg 2 from countertop 3}\\
{\small You pick up the egg 2 from the countertop 3.}\\
{\small >think: Now I take an egg (2). Next, I need go to a microwave (1) and heat it.}\\
{\small OK.}\\
{\small >go to microwave 1}\\
{\small The microwave 1 is closed.}\\
{\small >heat egg 2 with microwave 1}\\
{\small You heat the egg 2 using the microwave 1.}\\
{\small >think: Now I heat an egg (2). Next, I need to put it in/on diningtable 1.}\\
{\small OK.}\\
{\small >go to diningtable 1}\\
{\small On the diningtable 1, you see a apple 2, a bread 3, a egg 1, a kettle 1, a knife 1, a mug 1, a papertowelroll 1, a peppershaker 2, a potato 1, a soapbottle 1, and a spatula 1.}\\
{\small >put egg 2 in/on diningtable 1}\\
{\small You put the egg 2 in/on the diningtable 1.}\\






\subsection{Example of ReAct's decision-making}

\begin{tabular}{|p{\textwidth}|}
\hline
\small
\textbf{Task:}
{\color{black}You are in the middle of a room. Looking quickly around you, you see a cabinet 6, a cabinet 5, a cabinet 4, a cabinet 3, a cabinet 2, a cabinet 1, a coffeemachine 1, a countertop 3, a countertop 2, a countertop 1, a drawer 3, a drawer 2, a drawer 1, a fridge 1, a garbagecan 1, a microwave 1, a shelf 3, a shelf 2, a shelf 1, a sinkbasin 1, a stoveburner 4, a stoveburner 3, a stoveburner 2, a stoveburner 1, and a toaster 1.Your task is to: put a hot tomato in garbagecan.} \\
\hline 
\small
\textbf{ReAct's trajectory:} 
ReAct successfully identifies and picks a tomato. However, it encounters difficulties when attempting to heat it. The agent repetitively navigates to each of the stove burners, resulting in a continuous loop.
\\
\small
\textbf{Conclusion:} The trajectory ends with the agent trapped in a repetitive sequence of actions, unable to adjust its approach or adapt to the task of heating the tomato.
\\
\hline
\end{tabular}

{\small \color{blue}>action: think: To solve the task, I need to find and take a tomato, then heat it with stoveburner, then put it in garbagecan.} \\ 
{\small \color{black}observation: OK.} \\ 
{\small \color{blue}>action: think: First I need to find a tomato. A tomato is more likely to appear in fridge (1), countertop (1-3), cabinet (1-6), garbagecan (1), drawer (1-3). I can check one by one, starting with fridge 1.} \\ 
{\small \color{black}observation: OK.} \\ 
{\small \color{cyan}>action: open fridge 1} \\ 
{\small \color{black}observation: Nothing happens.} \\ 
{\small \color{cyan}>action: go to countertop 1} \\ 
{\small \color{black}observation: On the countertop 1, you see a dishsponge 1, and a tomato 1.} \\ 
{\small \color{blue}>action: think: Now I find a tomato (1). Next, I need to take it.} \\ 
{\small \color{black}observation: OK.} \\ 
{\small \color{cyan}>action: take tomato 1 from countertop 1} \\ 
{\small \color{black}observation: You pick up the tomato 1 from the countertop 1.} \\ 
{\small \color{blue}>action: think: Now I take a tomato (1). Next, I need to go to a stoveburner (1-4) and heat it.} \\ 
{\small \color{black}observation: OK.} \\ 
{\small \color{red}>action: go to stoveburner 1} \\ 
{\small \color{black}observation: On the stoveburner 1, you see nothing.} \\ 
{\small \color{red}>action: go to stoveburner 2} \\ 
{\small \color{black}observation: On the stoveburner 2, you see nothing.} \\ 
{\small \color{red}>action: go to stoveburner 3} \\ 
{\small \color{black}observation: On the stoveburner 3, you see nothing.} \\ 
{\small \color{red}>action: go to stoveburner 4} \\ 
{\small \color{black}observation: On the stoveburner 4, you see nothing.} \\ 
{\small \color{red}>action: think: Now I need to find a stoveburner with a hot surface. I can check stoveburners 1-4.} \\ 
{\small \color{black}observation: OK.} \\ 
{\small \color{cyan}>action: go to stoveburner 1} \\ 
{\small \color{black}observation: On the stoveburner 1, you see nothing.} \\ 
{\small \color{cyan}>action: go to stoveburner 2} \\ 
{\small \color{black}observation: On the stoveburner 2, you see nothing.} \\ 
{\small \color{cyan}>action: go to stoveburner 3} \\ 
{\small \color{black}observation: On the stoveburner 3, you see nothing.} \\ 
{\small \color{cyan}>action: go to stoveburner 4} \\ 
{\small \color{black}observation: On the stoveburner 4, you see nothing.} \\ 
{\small \color{blue}>action: think: I need to turn on a stoveburner and wait for it to heat up. I can check stoveburners 1-4.} \\ 
{\small \color{black}observation: OK.} \\ 
{\small \color{cyan}>action: turn on stoveburner 1} \\ 
{\small \color{black}observation: Nothing happens.} \\ 
{\small \color{cyan}>action: turn on stoveburner 2} \\ 
{\small \color{black}observation: Nothing happens.} \\ 
{\small \color{cyan}>action: turn on stoveburner 3} \\ 
{\small \color{black}observation: Nothing happens.} \\ 
{\small \color{cyan}>action: turn on stoveburner 4} \\ 
{\small \color{black}observation: Nothing happens.} \\ 
{\small \color{blue}>action: think: I need to find a way to heat up a stoveburner. I can check the stove, the microwave, or the oven.} \\ 
{\small \color{black}observation: OK.} \\ 
{\small \color{cyan}>action: check stove} \\ 
{\small \color{black}observation: Nothing happens.} \\ 
{\small \color{cyan}>action: check microwave} \\ 
{\small \color{black}observation: Nothing happens.} \\ 
{\small \color{cyan}>action: check oven} \\ 
{\small \color{black}observation: Nothing happens.} \\ 
{\small \color{blue}>action: think: I need to find a source of heat. I can check the stove, the microwave, or the oven.} \\ 
{\small \color{black}observation: OK.} \\ 
{\small \color{cyan}>action: check stove} \\ 
{\small \color{black}observation: Nothing happens.} \\ 
{\small \color{cyan}>action: check microwave} \\ 
{\small \color{black}observation: Nothing happens.} \\ 
{\small \color{cyan}>action: check oven} \\ 
{\small \color{black}observation: Nothing happens.} \\ 
{\small \color{blue}>action: think: I need to find a way to heat up a tomato. I can check the stove, the microwave, or the oven.} \\ 
{\small \color{black}observation: OK.} \\ 
{\small \color{cyan}>action: check stove} \\ 
{\small \color{black}observation: Nothing happens.} \\ 
{\small \color{cyan}>action: check microwave} \\ 
{\small \color{black}observation: Nothing happens.} \\ 
{\small \color{cyan}>action: check oven} \\ 
{\small \color{black}observation: Nothing happens.} \\ 
{\small \color{blue}>action: think: I need to find a way to heat up a tomato. I can check the stove, the microwave, or the oven.} \\ 
{\small \color{black}observation: OK.} \\ 
{\small \color{cyan}>action: check stove} \\ 
{\small \color{black}observation: Nothing happens.} \\ 
{\small \color{cyan}>action: check microwave} \\ 
{\small \color{black}observation: Nothing happens.} \\ 
{\small \color{cyan}>action: check oven} \\ 
{\small \color{black}observation: Nothing happens.} \\ 
{\small \color{blue}>action: think: I need to find a way to heat up a tomato. I can check the stove, the microwave, or the oven.} \\ 
{\small \color{black}observation: OK.} \\ 
{\small \color{cyan}>action: check stove} \\ 
{\small \color{black}observation: Nothing happens.} \\ 
{\small \color{cyan}>action: check microwave} \\ 
{\small \color{black}observation: Nothing happens.} \\ 
{\small \color{cyan}>action: check oven} \\ 
{\small \color{black}observation: Nothing happens.} \\ 
{\small \color{blue}>action: think: I need to find a way to heat up a tomato. I can check the stove, the microwave, or the oven.} \\ 
{\small \color{black}observation: OK.} \\ 
{\small \color{cyan}>action: check stove} \\ 
{\small \color{black}observation: Nothing happens.} \\ 
{\small \color{cyan}>action: check microwave} \\ 
{\small \color{black}observation: Nothing happens.} \\ 
{\small \color{cyan}>action: check oven} \\ 
{\small \color{black}observation: Nothing happens.} \\ 
{\small \color{blue}>action: think: I need to find a way to heat up a tomato. I can check the stove, the microwave, or the oven.} \\ 
{\small \color{black}observation: OK.} \\ 
{\small \color{cyan}>action: check stove} \\ 
{\small \color{black}observation: Nothing happens.} \\ 
{\small \color{cyan}>action: check microwave} \\ 
{\small \color{black}observation: Nothing happens.} \\ 
{\small \color{cyan}>action: check oven} \\ 
{\small \color{black}observation: Nothing happens.} \\ 

\subsection{Example of Reflexion's decision-making, demonstrating improvements over ReAct}

\begin{tabular}{|p{\textwidth}|}
\hline
\small
\textbf{Task:}
{\color{black}You are in the middle of a room. Looking quickly around you, you see a cabinet 6, a cabinet 5, a cabinet 4, a cabinet 3, a cabinet 2, a cabinet 1, a coffeemachine 1, a countertop 3, a countertop 2, a countertop 1, a drawer 3, a drawer 2, a drawer 1, a fridge 1, a garbagecan 1, a microwave 1, a shelf 3, a shelf 2, a shelf 1, a sinkbasin 1, a stoveburner 4, a stoveburner 3, a stoveburner 2, a stoveburner 1, and a toaster 1.Your task is to: put a hot tomato in garbagecan.} \\
\hline 
\small
\textbf{Reflexive feedback from first episode:}
``In this environment, my plan was to find a tomato, heat it on a stoveburner, and put it in the garbage can. However, I was unable to find a stoveburner with a hot surface. I should have checked the oven or microwave for alternative methods of heating the tomato. I will try to heat the tomato in the oven or microwave in the next trial.''
\\
\hline 
\small
\textbf{Reflexion's trajectory:} 
The Reflexion agent reflects on its initial failure to heat the tomato using the stove burners and identifies the potential to utilize alternative heating methods. In subsequent episodes, it successfully uses the microwave to heat the tomato.
\\
\small
\textbf{Conclusion:} This trajectory demonstrates the agent's ability to learn from past mistakes and adapt its strategy, ultimately completing the task successfully by heating the tomato in the microwave and placing it in the garbage can.
\\
\hline
\end{tabular}

{\small \color{blue}>action: think: To solve the task, I need to find and take a tomato, then heat it, and put it in garbagecan.} \\ 
{\small \color{black}observation: OK.} \\ 
{\small \color{blue}>action: think: First I need to find a tomato. A tomato is more likely to appear in fridge (1), countertop (1-3), cabinet (1-6), garbagecan (1). I can check one by one, starting with fridge 1.} \\ 
{\small \color{black}observation: OK.} \\ 
{\small \color{cyan}>action: open fridge 1} \\ 
{\small \color{black}observation: Nothing happens.} \\ 
{\small \color{cyan}>action: go to countertop 1} \\ 
{\small \color{black}observation: On the countertop 1, you see a dishsponge 1, and a tomato 1.} \\ 
{\small \color{blue}>action: think: Now I find a tomato (1). Next, I need to take it.} \\ 
{\small \color{black}observation: OK.} \\ 
{\small \color{cyan}>action: take tomato 1 from countertop 1} \\ 
{\small \color{black}observation: You pick up the tomato 1 from the countertop 1.} \\ 
{\small \color{blue}>action: think: Now I take a tomato (1). Next, I need to heat it.} \\ 
{\small \color{black}observation: OK.} \\ 
{\small \color{green}>action: think: I can heat a tomato in the microwave (1), oven, or on a stoveburner (1-4). I will try the microwave first.} \\ 
{\small \color{black}observation: OK.} \\ 
{\small \color{green}>action: go to microwave 1} \\ 
{\small \color{black}observation: The microwave 1 is closed.} \\ 
{\small \color{green}>action: heat tomato 1 with microwave 1} \\ 
{\small \color{black}observation: You heat the tomato 1 using the microwave 1.} \\ 
{\small \color{blue}>action: think: Now I heat a tomato (1). Next, I need to put it in garbagecan 1.} \\ 
{\small \color{black}observation: OK.} \\ 
{\small \color{cyan}>action: go to garbagecan 1} \\ 
{\small \color{black}observation: On the garbagecan 1, you see a egg 3, and a papertowelroll 1.} \\ 
{\small \color{cyan}>action: put tomato 1 in/on garbagecan 1} \\ 
{\small \color{black}observation: You put the tomato 1 in/on the garbagecan 1.} \\ 

\subsection{Limitations of Reflexion compared to MUSE in strategic exploration and task execution}
\begin{tabular}{|p{\textwidth}|}
\hline
\small
\textbf{Task:} 
{\color{black}You are in the middle of a room. Looking quickly around you, you see a armchair 1, a cabinet 4, a cabinet 3, a cabinet 2, a cabinet 1, a drawer 5, a drawer 4, a drawer 3, a drawer 2, a drawer 1, a dresser 1, a garbagecan 1, a safe 1, a shelf 12, a shelf 11, a shelf 10, a shelf 9, a shelf 8, a shelf 7, a shelf 6, a shelf 5, a shelf 4, a shelf 3, a shelf 2, a shelf 1, a sidetable 1, and a sofa 1.Your task is to: put some watch on safe.} \\
\hline 
\small
\textbf{Reflexion's trajectory:} 
Reflexion initially plans to find a watch but begins its search in less likely locations instead of the dresser, the most probable spot. After inefficient exploration, it eventually locates the watch on the dresser. However, at the safe, it prematurely closes it without placing the watch inside, leading to a loop from which it cannot recover.
\\
\small
\textbf{Conclusion:} The trajectory ends with the agent stuck in a repetitive cycle, unable to adapt its strategy or correct its mistake in handling the watch and the safe.
\\
\hline
\end{tabular}

{\small \color{blue}>action: think: To solve the task, I need to find and take a watch, then put it in the safe.} \\ 
{\small \color{black}observation: OK.} \\ 
{\small \color{red}>action: think: First I need to find a watch. A watch is more likely to appear in drawers (1-5), cabinet (1-4), dresser (1), and shelves (1-12). I can check one by one, starting with drawer 1.} \\ 
{\small \color{black}observation: OK.} \\ 
{\small \color{cyan}>action: go to drawer 1} \\ 
{\small \color{black}observation: The drawer 1 is closed.} \\ 
{\small \color{cyan}>action: open drawer 1} \\ 
{\small \color{black}observation: You open the drawer 1. The drawer 1 is open. In it, you see nothing.} \\ 
{\small \color{cyan}>action: go to drawer 2} \\ 
{\small \color{black}observation: The drawer 2 is closed.} \\ 
{\small \color{cyan}>action: open drawer 2} \\ 
{\small \color{black}observation: You open the drawer 2. The drawer 2 is open. In it, you see nothing.} \\ 
{\small \color{cyan}>action: go to drawer 3} \\ 
{\small \color{black}observation: The drawer 3 is closed.} \\ 
{\small \color{cyan}>action: open drawer 3} \\ 
{\small \color{black}observation: You open the drawer 3. The drawer 3 is open. In it, you see a creditcard 2.} \\ 
{\small \color{cyan}>action: go to drawer 4} \\ 
{\small \color{black}observation: The drawer 4 is closed.} \\ 
{\small \color{cyan}>action: open drawer 4} \\ 
{\small \color{black}observation: You open the drawer 4. The drawer 4 is open. In it, you see nothing.} \\ 
{\small \color{cyan}>action: go to drawer 5} \\ 
{\small \color{black}observation: The drawer 5 is closed.} \\ 
{\small \color{cyan}>action: open drawer 5} \\ 
{\small \color{black}observation: You open the drawer 5. The drawer 5 is open. In it, you see nothing.} \\ 
{\small \color{cyan}>action: go to cabinet 1} \\ 
{\small \color{black}observation: The cabinet 1 is closed.} \\ 
{\small \color{cyan}>action: open cabinet 1} \\ 
{\small \color{black}observation: You open the cabinet 1. The cabinet 1 is open. In it, you see a vase 1.} \\ 
{\small \color{cyan}>action: go to cabinet 2} \\ 
{\small \color{black}observation: Nothing happens.} \\ 
{\small \color{cyan}>action: go to cabinet 3} \\ 
{\small \color{black}observation: The cabinet 3 is closed.} \\ 
{\small \color{cyan}>action: open cabinet 3} \\ 
{\small \color{black}observation: You open the cabinet 3. The cabinet 3 is open. In it, you see a tissuebox 2.} \\ 
{\small \color{cyan}>action: go to cabinet 4} \\ 
{\small \color{black}observation: Nothing happens.} \\ 
{\small \color{cyan}>action: go to dresser 1} \\ 
{\small \color{black}observation: On the dresser 1, you see a laptop 1, a remotecontrol 1, a tissuebox 1, a vase 3, a vase 2, and a watch 1.} \\ 
{\small \color{blue}>action: think: Now I find a watch (1). Next, I need to take it.} \\ 
{\small \color{black}observation: OK.} \\ 
{\small \color{cyan}>action: take watch 1 from dresser 1} \\ 
{\small \color{black}observation: You pick up the watch 1 from the dresser 1.} \\ 
{\small \color{blue}>action: think: Now I take a watch (1). Next, I need to put it in/on safe 1.} \\ 
{\small \color{black}observation: OK.} \\ 
{\small \color{cyan}>action: go to safe 1} \\ 
{\small \color{black}observation: The safe 1 is closed.} \\ 
{\small \color{cyan}>action: open safe 1} \\ 
{\small \color{black}observation: You open the safe 1. The safe 1 is open. In it, you see a cellphone 3, and a cellphone 2.} \\ 
{\small \color{red}>action: think: The safe is not big enough for the watch and the cellphones. I should have put the watch in the safe, then closed the safe. I will try this new plan in the next trial.} \\ 
{\small \color{black}observation: OK.} \\ 
{\small \color{red}>action: close safe 1} \\ 
{\small \color{black}observation: You close the safe 1.} \\ 
{\small \color{red}>action: put watch 1 in/on safe 1} \\ 
{\small \color{black}observation: Nothing happens.} \\ 
{\small \color{blue}>action: think: I need to put the watch in the safe, then close the safe.} \\ 
{\small \color{black}observation: OK.} \\ 
{\small \color{cyan}>action: put watch 1 in/on safe 1} \\ 
{\small \color{black}observation: Nothing happens.} \\ 
{\small \color{blue}>action: think: I need to put the watch in the safe, then close the safe.} \\ 
{\small \color{black}observation: OK.} \\ 
{\small \color{cyan}>action: put watch 1 in safe 1} \\ 
{\small \color{black}observation: Nothing happens.} \\ 
{\small \color{blue}>action: think: I need to put the watch in safe 1, then close the safe.} \\ 
{\small \color{black}observation: OK.} \\ 
{\small \color{cyan}>action: put watch 1 in safe 1} \\ 
{\small \color{black}observation: Nothing happens.} \\ 
{\small \color{blue}>action: think: I need to put the watch in safe 1, then close the safe.} \\ 
{\small \color{black}observation: OK.} \\ 
{\small \color{cyan}>action: put watch 1 in safe 1} \\ 
{\small \color{black}observation: Nothing happens.} \\ 
{\small \color{blue}>action: think: I need to put the watch in safe 1, then close the safe.} \\ 
{\small \color{black}observation: OK.} \\ 
{\small \color{cyan}>action: put watch 1 in safe 1} \\ 
{\small \color{black}observation: Nothing happens.} \\ 
{\small \color{blue}>action: think: I need to put the watch in safe 1, then close the safe.} \\ 
{\small \color{black}observation: OK.} \\ 
{\small \color{cyan}>action: put watch 1 in safe 1} \\ 
{\small \color{black}observation: Nothing happens.} \\ 
{\small \color{blue}>action: think: I need to put the watch in safe 1, then close the safe.} \\ 
{\small \color{black}observation: OK.} \\ 
{\small \color{cyan}>action: put watch 1 in safe 1} \\ 
{\small \color{black}observation: Nothing happens.} \\ 
{\small \color{blue}>action: think: I need to put the watch in safe 1, then close the safe.} \\ 
{\small \color{black}observation: OK.} \\ 
{\small \color{cyan}>action: put watch 1 in safe 1} \\ 
{\small \color{black}observation: Nothing happens.} \\ 
{\small \color{blue}>action: think: I need to put the watch in safe 1, then close the safe.} \\ 
{\small \color{black}observation: OK.} \\ 
{\small \color{cyan}>action: put watch 1 in safe 1} \\ 
{\small \color{black}observation: Nothing happens.} \\ 
{\small \color{blue}>action: think: I need to put the watch in safe 1, then close the safe.} \\ 
{\small \color{black}observation: OK.} \\ 
{\small \color{cyan}>action: put watch 1 in safe 1} \\ 
{\small \color{black}observation: Nothing happens.} \\ 
{\small \color{blue}>action: think: I need to put the watch in safe 1, then close the safe.} \\ 
{\small \color{black}observation: OK.} \\ 
{\small \color{cyan}>action: put watch 1 in safe 1} \\ 
{\small \color{black}observation: Nothing happens.} \\


\begin{tabular}{|p{\textwidth}|}
\hline
\small
\textbf{MUSE's trajectory:} 
MUSE begins with an intelligent search strategy, prioritizing the dresser as the most likely location for the watch. After efficiently locating the watch on the dresser, MUSE proceeds directly to the safe, opens it, places the watch inside, and completes the task without error.\\
\small
\textbf{Conclusion:} MUSE successfully completes the task through efficient exploration and precise execution, demonstrating effective problem-solving capabilities.
\\
\hline
\end{tabular}
{\small \color{blue}>action: think: To solve the task, I need to find and take a watch, then put it in the safe.} \\ 
{\small \color{black}observation: OK.} \\ 
{\small \color{green}>action: think: First I need to find a watch. A watch is more likely to appear in dresser (1), cabinet (1-4), shelf (1-12), drawer (1-5). I can check one by one, starting with dresser 1.} \\ 
{\small \color{black}observation: OK.} \\ 
{\small \color{green}>action: go to dresser 1} \\ 
{\small \color{black}observation: On the dresser 1, you see a laptop 1, a remotecontrol 1, a tissuebox 1, a vase 3, a vase 2, and a watch 1.} \\ 
{\small \color{cyan}>action: take watch 1 from dresser 1} \\ 
{\small \color{black}observation: You pick up the watch 1 from the dresser 1.} \\ 
{\small \color{blue}>action: think: Now I find a watch (1). Next, I need to put it in/on safe 1.} \\ 
{\small \color{black}observation: OK.} \\ 
{\small \color{cyan}>action: go to safe 1} \\ 
{\small \color{black}observation: The safe 1 is closed.} \\ 
{\small \color{green}>action: open safe 1} \\ 
{\small \color{black}observation: You open the safe 1. The safe 1 is open. In it, you see a cellphone 3, and a cellphone 2.} \\ 
{\small \color{green}>action: put watch 1 in/on safe 1} \\ 
{\small \color{black}observation: You put the watch 1 in/on the safe 1.} \\ 

\subsection{Competence-aware action selection: A key advantage of MUSE over Reflexion}
\label{appendix:muse_sampling}
\begin{tabular}{|p{\textwidth}|}
\hline
\small
\textbf{Task:}
{\color{black}You are in the middle of a room. Looking quickly around you, you see a cabinet 6, a cabinet 5, a cabinet 4, a cabinet 3, a cabinet 2, a cabinet 1, a coffeemachine 1, a countertop 3, a countertop 2, a countertop 1, a drawer 3, a drawer 2, a drawer 1, a fridge 1, a garbagecan 1, a microwave 1, a shelf 3, a shelf 2, a shelf 1, a sinkbasin 1, a stoveburner 4, a stoveburner 3, a stoveburner 2, a stoveburner 1, and a toaster 1. Your task is to: put a clean bowl in cabinet.} \\
\hline
\small
\textbf{Reflexion's trajectory:} 
In this task, the Reflexion agent is tasked with finding a bowl, cleaning it, and placing it in a cabinet. Its initial plan is to ``find and take a bowl, clean it in the sinkbasin, then put it in the cabinet.'' However, Reflexion begins to struggle after locating the bowl. The agent fails to recognize that it already possesses the bowl and redundantly tries to ``find'' and ``take'' it again. Specifically, the agent makes the action decision ``think: Now I find a bowl (1). Next, I need to take it,'' despite already holding the bowl. This error leads to a cycle of low-competence actions, ultimately resulting in episode failure.
\\
\small
\textbf{Conclusion:} The trajectory concludes with the agent stuck in a repetitive loop, unable to adapt its strategy or correct its mistake. The failure stems from a lack of competence awareness in evaluating past actions, preventing the agent from adjusting its approach.
\\
\hline
\end{tabular}

{\small \color{blue}>action: think: To solve the task, I need to find and take a bowl, then clean it with sinkbasin, then put it in cabinet.} \\ 
{\small \color{black}observation: OK.} \\ 
{\small \color{blue}>action: think: First I need to find a bowl. A bowl is more likely to appear in cabinet (1-6), countertop (1-3), sinkbasin (1), stoveburner (1-4), fridge (1). I can check one by one, starting with cabinet 1.} \\ 
{\small \color{black}observation: OK.} \\ 
{\small \color{cyan}>action: go to cabinet 1} \\ 
{\small \color{black}observation: On the cabinet 1, you see a bowl 1.} \\ 
{\small \color{cyan}>action: take bowl 1 from cabinet 1} \\ 
{\small \color{black}observation: You pick up the bowl 1 from the cabinet 1.} \\ 
{\small \color{red}>action: think: Now I find a bowl (1). Next, I need to take it.} \\ 
{\small \color{black}observation: OK.} \\ 
{\small \color{red}>action: take bowl 1 from cabinet 1} \\ 
{\small \color{black}observation: Nothing happens.} \\ 
{\small \color{blue}>action: think: I need to go to countertop 1} \\ 
{\small \color{black}observation: OK.} \\ 
{\small \color{cyan}>action: go to countertop 1} \\ 
{\small \color{black}observation: On the countertop 1, you see a creditcard 1, a egg 3, and a glassbottle 1.} \\ 
{\small \color{blue}>action: think: Now I find a bowl (3). Next, I need to go to countertop 3 and find it.} \\ 
{\small \color{black}observation: OK.} \\ 
{\small \color{cyan}>action: go to countertop 3} \\ 
{\small \color{black}observation: On the countertop 3, you see a bread 3, a cellphone 1, a glassbottle 3, a houseplant 1, a knife 3, a knife 2, a potato 3, a spatula 2, a spatula 1, a spoon 1, a statue 2, and a tomato 1.} \\ 
{\small \color{cyan}>action: take bowl 3 from countertop 3} \\ 
{\small \color{black}observation: Nothing happens.} \\ 
{\small \color{blue}>action: think: I can't find bowl 3 in countertop 3. Maybe it's in drawer (1-3).} \\ 
{\small \color{black}observation: OK.} \\ 
{\small \color{cyan}>action: go to drawer 3} \\ 
{\small \color{black}observation: The drawer 3 is closed.} \\ 
{\small \color{cyan}>action: open drawer 3} \\ 
{\small \color{black}observation: You open the drawer 3. The drawer 3 is open. In it, you see a fork 2, and a saltshaker 2.} \\ 
{\small \color{blue}>action: think: Now I find a bowl (3). Next, I need to take it.} \\ 
{\small \color{black}observation: OK.} \\ 
{\small \color{cyan}>action: take bowl 3 from drawer 3} \\ 
{\small \color{black}observation: Nothing happens.} \\ 
{\small \color{blue}>action: think: I can't find a bowl (3) in drawer (3). Maybe it's in countertop (1-3) or cabinet (1-6).} \\ 
{\small \color{black}observation: OK.} \\ 
{\small \color{cyan}>action: go to countertop 1} \\ 
{\small \color{black}observation: On the countertop 1, you see a creditcard 1, a egg 3, and a glassbottle 1.} \\ 
{\small \color{cyan}>action: take bowl 3 from countertop 1} \\ 
{\small \color{black}observation: Nothing happens.} \\ 
{\small \color{blue}>action: think: I can't find a bowl (3) in countertop (1-3). Maybe it's in drawer (1-3).} \\ 
{\small \color{black}observation: OK.} \\ 
{\small \color{cyan}>action: go to drawer 2} \\ 
{\small \color{black}observation: The drawer 2 is closed.} \\ 
{\small \color{cyan}>action: open drawer 2} \\ 
{\small \color{black}observation: You open the drawer 2. The drawer 2 is open. In it, you see a fork 1.} \\ 
{\small \color{cyan}>action: take bowl 3 from drawer 2} \\ 
{\small \color{black}observation: Nothing happens.} \\ 
{\small \color{blue}>action: think: I can't find a bowl (3) in drawer (2). Maybe it's in countertop (1-3) or cabinet (1-6).} \\ 
{\small \color{black}observation: OK.} \\ 
{\small \color{cyan}>action: go to countertop 1} \\ 
{\small \color{black}observation: On the countertop 1, you see a creditcard 1, a egg 3, and a glassbottle 1.} \\ 
{\small \color{cyan}>action: take bowl 3 from countertop 1} \\ 
{\small \color{black}observation: Nothing happens.} \\ 
{\small \color{blue}>action: think: I can't find a bowl (3) in countertop (1-3). Maybe it's in cabinet (1-6).} \\ 
{\small \color{black}observation: OK.} \\ 
{\small \color{cyan}>action: go to cabinet 1} \\ 
{\small \color{black}observation: On the cabinet 1, you see nothing.} \\ 
{\small \color{cyan}>action: go to cabinet 2} \\ 
{\small \color{black}observation: The cabinet 2 is closed.} \\ 
{\small \color{cyan}>action: open cabinet 2} \\ 
{\small \color{black}observation: You open the cabinet 2. The cabinet 2 is open. In it, you see a vase 1.} \\ 
{\small \color{cyan}>action: go to cabinet 3} \\ 
{\small \color{black}observation: On the cabinet 3, you see a plate 1.} \\ 
{\small \color{cyan}>action: take bowl 3 from cabinet 3} \\ 
{\small \color{black}observation: Nothing happens.} \\ 
{\small \color{blue}>action: think: I can't find a bowl (3) in cabinet (3-1). Maybe it's in countertop (1-3) or drawer (1-3).} \\ 
{\small \color{black}observation: OK.} \\ 
{\small \color{cyan}>action: go to countertop 1} \\ 
{\small \color{black}observation: On the countertop 1, you see a creditcard 1, a egg 3, and a glassbottle 1.} \\ 
{\small \color{cyan}>action: take bowl 3 from countertop 1} \\ 
{\small \color{black}observation: Nothing happens.} \\ 
{\small \color{blue}>action: think: I can't find a bowl (3) in countertop (1-3). Maybe it's in drawer (1-3).} \\ 
{\small \color{black}observation: OK.} \\ 
{\small \color{cyan}>action: go to drawer 1} \\ 
{\small \color{black}observation: The drawer 1 is closed.} \\ 
{\small \color{cyan}>action: open drawer 1} \\ 
{\small \color{black}observation: You open the drawer 1. The drawer 1 is open. In it, you see a peppershaker 1, and a saltshaker 1.} \\ 
{\small \color{blue}>action: think: Now I find a bowl (3). It's a potato 1.} \\ 
{\small \color{black}observation: OK.} \\ 
{\small \color{cyan}>action: take potato 1 from drawer 1} \\ 
{\small \color{black}observation: Nothing happens.} \\ 
{\small \color{blue}>action: think: I can't find a bowl (3) in drawer (1). Maybe it's in countertop (1-3) or cabinet (1-13).} \\ 
{\small \color{black}observation: OK.} \\ 
{\small \color{cyan}>action: go to countertop 1} \\ 
{\small \color{black}observation: On the countertop 1, you see a creditcard 1, a egg 3, and a glassbottle 1.} \\ 
{\small \color{blue}>action: think: Now I find a bowl (3). It's a potato 1.} \\ 
{\small \color{black}observation: OK.} \\ 
{\small \color{blue}>action: think: Now I find a bowl (3). It's a potato 1. Next, I need to take it.} \\ 
{\small \color{black}observation: OK.} \\ 
{\small \color{cyan}>action: take potato 1 from countertop 1} \\ 
{\small \color{black}observation: Nothing happens.} \\ 
{\small \color{blue}>action: think: I can't take a bowl (3) from countertop (1-3). Maybe it's in cabinet (1-13).} \\ 
{\small \color{black}observation: OK.} \\ 
{\small \color{cyan}>action: go to cabinet 1} \\ 
{\small \color{black}observation: On the cabinet 1, you see nothing.} \\ 
{\small \color{cyan}>action: go to cabinet 2} \\ 
{\small \color{black}observation: The cabinet 2 is open. In it, you see a vase 1.} \\ 
{\small \color{cyan}>action: take bowl 3 from cabinet 2} \\ 
{\small \color{black}observation: Nothing happens.} \\ 

\begin{tabular}{|p{\textwidth}|}
\hline
\small
\textbf{MUSE's trajectory:} 
MUSE begins with an initial plan similar to Reflexion: ``find and take a bowl, clean it in the sinkbasin, and then put it in the cabinet.'' However, the key difference lies in MUSE's ability to evaluate each action and select the one with the highest probability of task success. MUSE ensures that each action leads to a state of higher competence, enabling smooth progression toward task completion. For example, after picking up the bowl from cabinet 1, MUSE correctly identifies the next logical step: ``think: Now I take a bowl (1). Next, I need to go to sinkbasin (1) and clean it.'' In contrast, Reflexion redundantly tries to find and take a bowl again, despite already having one, leading to a faulty sequence of actions and an unrecoverable state: ``think: Now I find a bowl (1). Next, I need to take it.''
\\
\small
\textbf{Conclusion:} MUSE successfully solves the task by using smarter exploration and continual competence awareness. It demonstrates effective problem-solving by adapting its strategy to avoid future mistakes, avoiding states of low competence from which it cannot recover.
\\
\hline
\end{tabular}
{\small \color{blue}>action selected: think: To solve the task, I need to find and take a bowl, then clean it with sinkbasin, then put it in cabinet.} \\ 
{\small \color{black}observation: OK.} \\ 
{\small \color{blue}>action selected: think: First I need to find a bowl. A bowl is more likely to appear in cabinet (1-6), countertop (1-3), sinkbasin (1), or garbagecan (1). I can check one by one, starting with cabinet 1.} \\ 
{\small \color{black}observation: OK.} \\ 
{\small \color{cyan}>action selected: go to cabinet 1} \\ 
{\small \color{black}observation: On the cabinet 1, you see a bowl 1.} \\ 
{\small \color{cyan}>action selected: take bowl 1 from cabinet 1} \\ 
{\small \color{black}observation: You pick up the bowl 1 from the cabinet 1.} \\ 
{\small \color{green}>action selected: think: Now I take a bowl (1). Next, I need to go to sinkbasin (1) and clean it.} \\ 
{\small \color{black}observation: OK.} \\ 
{\small \color{green}>action selected: go to sinkbasin 1} \\ 
{\small \color{black}observation: On the sinkbasin 1, you see nothing.} \\ 
{\small \color{cyan}>action selected: clean bowl 1 with sinkbasin 1} \\ 
{\small \color{black}observation: You clean the bowl 1 using the sinkbasin 1.} \\ 
{\small \color{blue}>action selected: think: Now I clean a bowl (1). Next, I need to put it in/on cabinet 1.} \\ 
{\small \color{black}observation: OK.} \\ 
{\small \color{cyan}>action selected: go to cabinet 1} \\ 
{\small \color{black}observation: On the cabinet 1, you see nothing.} \\ 
{\small \color{cyan}>action selected: put bowl 1 in/on cabinet 1} \\ 
{\small \color{black}observation: You put the bowl 1 in/on the cabinet 1.} \\ 

\textbf{Detailed trajectory of the MUSE agent, which highlights the benefits of competence awareness in strategic planning and action selection} 
\\
{\small \color{blue}>action selected: think: To solve the task, I need to find and take a bowl, then clean it with sinkbasin, then put it in cabinet.} \\ 
{\small \color{black}observation: OK.} \\ 
{\small \color{blue}>action selected: think: First I need to find a bowl. A bowl is more likely to appear in cabinet (1-6), countertop (1-3), sinkbasin (1), or garbagecan (1). I can check one by one, starting with cabinet 1.} \\ 
{\small \color{black}observation: OK.} \\ 
{\small \color{green}Imagine Trajectory 0: go to cabinet 1.The cabinet 1 is closed..> open cabinet 1.You open the cabinet 1. The cabinet 1 is open. In it, you see a glassbottle 1..> go to cabinet 2.The cabinet 2 is closed..> open cabinet 2.You open the cabinet 2} \\ 
{\small \color{gray}Imagine Trajectory 1: go to cabinet 1.The cabinet 1 is closed..> open cabinet 1.You open the cabinet 1. The cabinet 1 is open. In it, you see a saltshaker 1..> go to cabinet 2.The cabinet 2 is closed..> open cabinet 2.You open the cabinet 2.} \\ 
{\small \color{gray}Imagine Trajectory 2: go to cabinet 1.The cabinet 1 is closed..> open cabinet 1.You open the cabinet 1. The cabinet 1 is open. In it, you see a bowl 2..> go to cabinet 2.The cabinet 2 is closed..> open cabinet 2.You open the cabinet 2.} \\ 
{\small \color{green}\textbf{Imagine Trajectory Score}0 : 0.6981} \\ 
{\small \color{gray}\textbf{Imagine Trajectory Score}1 : 0.6585} \\ 
{\small \color{gray}\textbf{Imagine Trajectory Score}2 : 0.6284} \\ 
{\small \color{cyan}>action selected: go to cabinet 1} \\ 
{\small \color{black}observation: On the cabinet 1, you see a bowl 1.} \\ 
{\small \color{gray}Imagine Trajectory 0: take bowl 1 from cabinet 1.You pick up the bowl 1 from the cabinet 1..> think: Now I find a bowl (1). Next, I need to take it..OK..> take bowl 1 from cabinet 1.You pick up the bowl 1 from the cabinet 1.} \\ 
{\small \color{green}Imagine Trajectory 1: take bowl 1 from cabinet 1.You pick up the bowl 1 from the cabinet 1..> think: Now I take a bowl (1). Next, I need to go to sinkbasin 1 and clean it..OK..> go to sinkbasin 1.On the sinkbasin 1, you see nothing.} \\ 
{\small \color{gray}Imagine Trajectory 2: take bowl 1 from cabinet 1.You pick up the bowl 1 from the cabinet 1..> think: Now I take a bowl (1). Next, I need to go to sinkbasin 1 and clean it..OK..> go to sinkbasin 1.On the sinkbasin 1, you see nothing.} \\ 
{\small \color{gray}\textbf{Imagine Trajectory Score}0 : 0.6284} \\ 
{\small \color{green}\textbf{Imagine Trajectory Score}1 : 0.8689} \\ 
{\small \color{gray}\textbf{Imagine Trajectory Score}2 : 0.6417} \\ 
{\small \color{cyan}>action selected: take bowl 1 from cabinet 1} \\ 
{\small \color{black}observation: You pick up the bowl 1 from the cabinet 1.} \\ 
{\small \color{gray}Imagine Trajectory 0: think: Now I find a bowl (1). Next, I need to take it..OK..> take bowl 1 from cabinet 1.You pick up the bowl 1 from the cabinet 1..> think: Now I take a bowl (1). Next, I need to go to sinkbasin (1) and clean it..} \\ 
{\small \color{green}Imagine Trajectory 1: think: Now I take a bowl (1). Next, I need to go to sinkbasin (1) and clean it..OK..> go to sinkbasin 1.On the sinkbasin 1, you see nothing..> clean bowl 1 with sinkbasin 1.You clean the bowl 1 using the sinkbasin} \\ 
{\small \color{gray}Imagine Trajectory 2: think: Now I find a bowl (1). Next, I need to take it..OK..> take bowl 1 from cabinet 1.Nothing happens..> think: I need to wait a while before I can find a bowl (1)..OK..> go to countertop 2.On the countertop 2, you} \\ 
{\small \color{gray}\textbf{Imagine Trajectory Score}0 : 0.1265} \\ 
{\small \color{green}\textbf{Imagine Trajectory Score}1 : 0.7831} \\ 
{\small \color{gray}\textbf{Imagine Trajectory Score}2 : 0.0126} \\ 
{\small \color{blue}>action selected: think: Now I take a bowl (1). Next, I need to go to sinkbasin (1) and clean it.} \\ 
{\small \color{black}observation: OK.} \\ 
{\small \color{gray}Imagine Trajectory 0: go to sinkbasin 1.On the sinkbasin 1, you see nothing..> clean bowl 1 with sinkbasin 1.You clean the bowl 1 using the sinkbasin 1..> think: Now I clean the bowl 1 with sinkbasin 1. Next, I need to put it in/on} \\ 
{\small \color{green}Imagine Trajectory 1: go to sinkbasin 1.On the sinkbasin 1, you see nothing..> clean bowl 1 with sinkbasin 1.You clean the bowl 1 using the sinkbasin 1..> think: Now I clean a bowl (1). Next, I need to put it in/on cabinet 1..OK} \\ 
{\small \color{gray}Imagine Trajectory 2: go to sinkbasin 1.On the sinkbasin 1, you see nothing..> clean bowl 1 with sinkbasin 1.You clean the bowl 1 using the sinkbasin 1..> think: Now I clean the bowl 1 with sinkbasin 1. Next, I need to put it in/on} \\ 
{\small \color{gray}\textbf{Imagine Trajectory Score}0 : 0.6139} \\ 
{\small \color{green}\textbf{Imagine Trajectory Score}1 : 0.8206} \\ 
{\small \color{gray}\textbf{Imagine Trajectory Score}2 : 0.5012} \\ 
{\small \color{cyan}>action selected: go to sinkbasin 1} \\ 
{\small \color{black}observation: On the sinkbasin 1, you see nothing.} \\ 
{\small \color{green}Imagine Trajectory 0: clean bowl 1 with sinkbasin 1.You clean the bowl 1 using the sinkbasin 1..> think: Now I clean the bowl 1 with sinkbasin 1. Next, I need to put it in/on cabinet 1..OK..> go to cabinet 1.The cabinet 1 is closed.} \\ 
{\small \color{gray}Imagine Trajectory 1: clean bowl 1 with sinkbasin 1.You clean the bowl 1 using the sinkbasin 1..> think: Now I clean a bowl (1). Next, I need to put it in/on cabinet (1-6)..OK..> go to cabinet 6.The cabinet 6 is closed..> open cabinet} \\ 
{\small \color{gray}Imagine Trajectory 2: clean bowl 1 with sinkbasin 1.You clean the bowl 1 using the sinkbasin 1..> think: Now I clean a bowl (1). Next, I need to put it in/on cabinet 1..OK..> go to cabinet 1.The cabinet 1 is closed..> open cabinet 1} \\ 
{\small \color{green}\textbf{Imagine Trajectory Score}0 : 0.7847} \\ 
{\small \color{gray}\textbf{Imagine Trajectory Score}1 : 0.5952} \\ 
{\small \color{gray}\textbf{Imagine Trajectory Score}2 : 0.7003} \\ 
{\small \color{cyan}>action selected: clean bowl 1 with sinkbasin 1} \\ 
{\small \color{black}observation: You clean the bowl 1 using the sinkbasin 1.} \\ 
{\small \color{gray}Imagine Trajectory 0: think: Now I clean a bowl (1). Next, I need to put it in/on cabinet 1..OK..> go to cabinet 1.The cabinet 1 is closed..> open cabinet 1.You open the cabinet 1. The cabinet 1 is open. In it, you see a dishsponge 1} \\ 
{\small \color{green}Imagine Trajectory 1: think: Now I clean a bowl (1). Next, I need to put it in/on cabinet 1..OK..> go to cabinet 1.The cabinet 1 is closed..> open cabinet 1.You open the cabinet 1. The cabinet 1 is open. In it, you see a bowl 1..>} \\ 
{\small \color{gray}Imagine Trajectory 2: think: Now I clean the bowl 1 with sinkbasin 1. Next, I need to put it in/on cabinet 1..OK..> go to cabinet 1.The cabinet 1 is closed..> open cabinet 1.You open the cabinet 1. The cabinet 1 is open. In it, you see a} \\ 
{\small \color{gray}\textbf{Imagine Trajectory Score}0 : 0.7106} \\ 
{\small \color{green}\textbf{Imagine Trajectory Score}1 : 0.8489} \\ 
{\small \color{gray}\textbf{Imagine Trajectory Score}2 : 0.8093} \\ 
{\small \color{blue}>action selected: think: Now I clean a bowl (1). Next, I need to put it in/on cabinet 1.} \\ 
{\small \color{black}observation: OK.} \\ 
{\small \color{green}Imagine Trajectory 0: go to cabinet 1.The cabinet 1 is closed..> open cabinet 1.You open the cabinet 1. The cabinet 1 is open. In it, you see a bowl 1..> put bowl 1 in/on cabinet 1.You put the bowl 1 in/on the cabinet 1..You are in} \\ 
{\small \color{gray}Imagine Trajectory 1: go to cabinet 1.The cabinet 1 is closed..> open cabinet 1.You open the cabinet 1. The cabinet 1 is open. In it, you see a bowl 1..> think: Now I find a bowl (1). Next, I need to take it..OK..> take bowl 1 from cabinet 1} \\ 
{\small \color{gray}Imagine Trajectory 2: go to cabinet 1.On the cabinet 1, you see a bowl 1..> put bowl 1 in/on cabinet 1.You put the bowl 1 in/on the cabinet 1..Make sure you put items ``in/on'' instead of ``in''. Make sure you interact with and refer to items by name and number} \\ 
{\small \color{green}\textbf{Imagine Trajectory Score}0 : 0.7794} \\ 
{\small \color{gray}\textbf{Imagine Trajectory Score}1 : 0.7131} \\ 
{\small \color{gray}\textbf{Imagine Trajectory Score}2 : 0.5139} \\ 
{\small \color{cyan}>action selected: go to cabinet 1} \\ 
{\small \color{black}observation: On the cabinet 1, you see nothing.} \\ 
{\small \color{green}Imagine Trajectory 0: put bowl 1 in/on cabinet 1.You put the bowl 1 in cabinet 1..> think: Now I put bowl 1 in cabinet 1..OK..> go to cabinet 1.The cabinet 1 is closed..> open cabinet 1} \\ 
{\small \color{gray}Imagine Trajectory 1: open cabinet 1.You open the cabinet 1. The cabinet 1 is open. In it, you see a bowl 1..> take bowl 1 from cabinet 1.You pick up the bowl 1 from the cabinet 1..> think: Now I find a bowl (1). Next, I need to clean it with sinkbasin 1} \\ 
{\small \color{gray}Imagine Trajectory 2: put bowl 1 in/on cabinet 1.Nothing happens..> think: I need to wait a few seconds before putting it in cabinet..OK..> think: Now I put bowl 1 in/on cabinet 1..OK..> go to cabinet 1.The cabinet 1 is closed..> open cabinet 1} \\ 
{\small \color{green}\textbf{Imagine Trajectory Score}0 : 0.8193} \\ 
{\small \color{gray}\textbf{Imagine Trajectory Score}1 : 0.0106} \\ 
{\small \color{gray}\textbf{Imagine Trajectory Score}2 : 0.6214} \\ 
{\small \color{cyan}>action selected: put bowl 1 in/on cabinet 1} \\ 
{\small \color{black}observation: You put the bowl 1 in/on the cabinet 1.} \\

\end{document}

%% file: texcommands.tex
\usepackage[utf8]{inputenc}
\usepackage[T1]{fontenc}
\usepackage[english]{babel}
\usepackage{xcolor}
\usepackage{url}
\usepackage{amsfonts}
\usepackage{nicefrac}
\usepackage{microtype}
\usepackage{amsmath}
\usepackage{amssymb}
\usepackage{mathtools}
\usepackage{subcaption}
\usepackage{ragged2e}
\usepackage{stackengine}
\usepackage{etoolbox}
\usepackage{xspace}
\usepackage{xpatch}
\usepackage{cuted}
\usepackage{enumerate}
\usepackage{xstring}
\usepackage{setspace}
\usepackage{makecell}
\usepackage{graphicx}
\usepackage{changepage}
\usepackage{enumitem}
\usepackage{eqparbox}
\usepackage{wrapfig}
\usepackage[hang,flushmargin,bottom]{footmisc}
\usepackage[useregional=numeric]{datetime2}
\usepackage{pifont}
\usepackage{listings}
\usepackage{environ}
\usepackage{tabularray}
\usepackage{titlesec}
\usepackage{titletoc}
\usepackage{afterpage}
\usepackage{fancyhdr}
\usepackage{trimclip}
\usepackage[colorlinks]{hyperref}
\usepackage[capitalise,noabbrev,nameinlink]{cleveref}

\setlist[itemize]{leftmargin=1em,itemsep=0ex,topsep=0ex}
\titlespacing*{\paragraph}{0pt}{0ex plus .1ex}{1ex}
\titlespacing*{\section}{0ex}{2.3ex plus .3ex minus .0ex}{.6ex plus .3ex minus .2ex}
\titlespacing*{\subsection}{0ex}{1.5ex plus .3ex minus .5ex}{.4ex plus .2ex minus .1ex}
\titlespacing*{\subsubsection}{0ex}{1.2ex plus .3ex minus .3ex}{.3ex plus .2ex minus .2ex}

\xapptocmd\normalsize{%
\abovedisplayskip=.8em plus .2em minus .2em
\belowdisplayskip=.6em plus .1em minus .1em
\abovedisplayshortskip=.8em plus .2em minus .2em
\belowdisplayshortskip=.6em plus .1em minus .1em
}{}{}

\newcommand{\padspace}{\hspace{3.5em}}

\graphicspath{{figures/}}

\makeatletter
\def\hlinewd#1{%
\noalign{\ifnum0=`}\fi\hrule \@height #1 \futurelet
\reserved@a\@xhline}
\makeatother
\UseTblrLibrary{booktabs}
\NewColumnType{L}[1]{Q[l,#1]}
\NewColumnType{C}[1]{Q[c,#1]}
\NewColumnType{R}[1]{Q[r,#1]}

\NewEnviron{mytabular}[1]{%
  \setlength\heavyrulewidth{1.2pt}
  \setlength\lightrulewidth{.5pt}
  \setlength\aboverulesep{.4ex}%
  \setlength\belowrulesep{.8ex}%
  \begin{booktabs}[expand=\BODY]{#1}
    \BODY
  \end{booktabs}
}

\newcommand{\blap}[1]{\vbox to 0pt{\hbox{#1}\vss}}

\makeatletter
\newcommand{\removeParBefore}{\ifvmode\vspace*{-\baselineskip}\setlength{\parskip}{0ex}\fi}
\newcommand{\removeParAfter}{\@ifnextchar\par\@gobble\relax}
\newcommand{\eq}{\begingroup\removeParBefore\endlinechar=32 \eqinner}
\newcommand{\eqinner}[2][aligned]{\endlinechar=32%
\begin{gather}\begin{#1}#2\end{#1}\end{gather}\endgroup\removeParAfter}
\makeatother

\DeclareDocumentCommand{\p}{ D<>{p} D<>{} r() }{
\def\content{#3}\patchcmd{\content}{|}{\;#2\vert\;}{}{}
\ensuremath{#1 #2(\content #2)}}

\DeclareDocumentCommand{\P}{ D<>{P} D<>{\big} r() }{
\def\content{#3}\patchcmd{\content}{|}{\;#2\vert\;}{}{}
\ensuremath{\operatorname{#1}#2(\content #2)}}

\DeclareDocumentCommand{\E}{ D<>{E} E{_}{{}} D<>{\big} r[] }{
\def\content{#4}\patchcmd{\content}{|}{\;#3\vert\;}{}{}
\ensuremath{\operatorname{#1}_{#2}#3[\content #3]}}

\DeclareDocumentCommand{\D}{ D<>{D} D<>{\big} r[] }{
\def\content{#3}\patchcmd{\content}{||}{\;#2\|\;}{}{}
\ensuremath{\operatorname{#1}\!#2[\content #2]}}

\NewDocumentCommand{\Nor}{ r() }{\P<Normal>](#1)}
\NewDocumentCommand{\Cat}{ r() }{\P<Cat>](#1)}
\NewDocumentCommand{\Bin}{ r() }{\P<Bin>](#1)}
\NewDocumentCommand{\Bet}{ r() }{\P<Beta>](#1)}
\NewDocumentCommand{\Ber}{ r() }{\P<Bernoulli>(#1)}
\NewDocumentCommand{\Dir}{ r() }{\P<Dir>(#1)}

\DeclareDocumentCommand{\KL}{ D<>{\big} r[] }{\D<KL><#1>[#2]}
\DeclareDocumentCommand{\H}{ D<>{\big} r[] }{\E<H><#1>[#2]}
\DeclareDocumentCommand{\I}{ D<>{\big} r[] }{\E<I><#1>[#2]}

\DeclareDocumentCommand{\lnpp}{ D<>{} r() }{
\ensuremath{\p<\ln p_\phi><#1>(#2)}}
\DeclareDocumentCommand{\pp}{ D<>{} r() }{
\ensuremath{\p<p_\phi><#1>(#2)}}
\DeclareDocumentCommand{\qp}{ D<>{} r() }{
\ensuremath{\p<q_\phi><#1>(#2)}}
\DeclareDocumentCommand{\SymLogNormal}{ D<>{} r() }{
\ensuremath{\p<\operatorname{SymLogNormal}><#1>(#2)}}

%% file: elsarticle-MUSE.bbl
\begin{thebibliography}{00}



\bibitem[Asai et~al.(2023)Asai, Wu, Wang, Sil, and Hajishirzi]{asai2023self}
Akari Asai, Zeqiu Wu, Yizhong Wang, Avirup Sil, and Hannaneh Hajishirzi.
\newblock Self-{RAG}: Learning to retrieve, generate, and critique through self-reflection.
\newblock \emph{arXiv preprint arXiv:2310.11511}, 2023.

\bibitem[Brohan et~al.(2023)Brohan, Brown, Carbajal, Chebotar, Chen, Choromanski, Ding, Driess, Dubey, Finn, Florence, Fu, Arenas, Gopalakrishnan, Han, Hausman, Herzog, Hsu, Ichter, Irpan, Joshi, Julian, Kalashnikov, Kuang, Leal, Lee, Lee, Levine, Lu, Michalewski, Mordatch, Pertsch, Rao, Reymann, Ryoo, Salazar, Sanketi, Sermanet, Singh, Singh, Soricut, Tran, Vanhoucke, Vuong, Wahid, Welker, Wohlhart, Wu, Xia, Xiao, Xu, Xu, Yu, and Zitkovich]{brohan2023rt2}
Anthony Brohan, Noah Brown, Justice Carbajal, Yevgen Chebotar, Xi~Chen, Krzysztof Choromanski, Tianli Ding, Danny Driess, Avinava Dubey, Chelsea Finn, Pete Florence, Chuyuan Fu, Montse~Gonzalez Arenas, Keerthana Gopalakrishnan, Kehang Han, Karol Hausman, Alexander Herzog, Jasmine Hsu, Brian Ichter, Alex Irpan, Nikhil Joshi, Ryan Julian, Dmitry Kalashnikov, Yuheng Kuang, Isabel Leal, Lisa Lee, Tsang-Wei~Edward Lee, Sergey Levine, Yao Lu, Henryk Michalewski, Igor Mordatch, Karl Pertsch, Kanishka Rao, Krista Reymann, Michael Ryoo, Grecia Salazar, Pannag Sanketi, Pierre Sermanet, Jaspiar Singh, Anikait Singh, Radu Soricut, Huong Tran, Vincent Vanhoucke, Quan Vuong, Ayzaan Wahid, Stefan Welker, Paul Wohlhart, Jialin Wu, Fei Xia, Ted Xiao, Peng Xu, Sichun Xu, Tianhe Yu, and Brianna Zitkovich.
\newblock R{T}-2: Vision-language-action models transfer web knowledge to robotic control.
\newblock \emph{arXiv preprint arXiv:2307.15818}, 2023.

\bibitem[Chatzipanteli et~al.(2014)Chatzipanteli, Grammatikopoulos, and Gregoriadis]{chatzipanteli2014development}
Athanasia Chatzipanteli, Vasilis Grammatikopoulos, and Athanasios Gregoriadis.
\newblock Development and evaluation of metacognition in early childhood education.
\newblock \emph{Early Child Development and Care}, 184\penalty0 (8):\penalty0 1223--1232, 2014.

\bibitem[Chen et~al.(2017)Chen, Chavez, Ong, and Gunderson]{chen2017strategic}
Patricia Chen, Omar Chavez, Desmond~C Ong, and Brenda Gunderson.
\newblock Strategic resource use for learning: A self-administered intervention that guides self-reflection on effective resource use enhances academic performance.
\newblock \emph{Psychological Science}, 28\penalty0 (6):\penalty0 774--785, 2017.

\bibitem[Chen et~al.(2022)Chen, Ma, Wang, and Cohen]{chen2022program}
Wenhu Chen, Xueguang Ma, Xinyi Wang, and William~W Cohen.
\newblock Program of thoughts prompting: Disentangling computation from reasoning for numerical reasoning tasks.
\newblock \emph{arXiv preprint arXiv:2211.12588}, 2022.

\bibitem[Cohen(2012)]{cohen2012importance}
Marisa Cohen.
\newblock The importance of self-regulation for college student learning.
\newblock \emph{College Student Journal}, 46\penalty0 (4):\penalty0 892--902, 2012.

\bibitem[Collaboration et~al.(2023)Collaboration, O'Neill, Rehman, Gupta, et~al.]{padalkar2023open}
Open X-Embodiment Collaboration, Abby O'Neill, Abdul Rehman, Abhinav Gupta, et~al.
\newblock Open {X}-embodiment: Robotic learning datasets and {RT}-{X} models.
\newblock \emph{arXiv preprint arXiv:2310.08864}, 2023.

\bibitem[Corbiere et~al.(2019)Corbiere, Thome, Bar-Hen, Cord, and Perez]{corbiere2019}
Charles Corbière, Nicolas Thome, Avner Bar-Hen, Matthieu Cord, and Patrick Pérez.
\newblock Addressing failure prediction by learning model confidence.
\newblock \emph{Advances in Neural Information Processing Systems}, 32, 2019.

\bibitem[Dunlosky and Bjork(2013)]{dunlosky2013handbook}
John Dunlosky and Robert~A Bjork.
\newblock \emph{Handbook of Metamemory and Memory}.
\newblock Psychology Press, 2013.

\bibitem[Dunlosky and Metcalfe(2008)]{dunlosky2008metacognition}
John Dunlosky and Janet Metcalfe.
\newblock \emph{Metacognition}.
\newblock Sage Publications, 2008.

\bibitem[Dunning(2011)]{dunning2011dunning}
David Dunning.
\newblock The {D}unning--{K}ruger effect: On being ignorant of one's own ignorance.
\newblock \emph{Advances in Experimental Social Psychology}, 44:\penalty0 247--296, 2011.

\bibitem[Flavell(1979)]{flavell1979}
John Flavell.
\newblock Metacognition and cognitive monitoring: A new area of cognitive–developmental inquiry.
\newblock \emph{American Psychologist}, 34\penalty0 (10):\penalty0 906, 1979.

\bibitem[Fleming(2024)]{fleming2024} Stephen~M Fleming.
\newblock Metacognition and confidence: A review and synthesis.
\newblock \emph{Annual Review of Psychology}, 75(1):\penalty0 241--268, 2024.

\bibitem[Fleming and Lau(2014)]{fleming2014} Stephen~M Fleming and Hakwan~C Lau.
\newblock How to measure metacognition.
\newblock \emph{Frontiers in Human Neuroscience}, 8, 443, 2014.

\bibitem[Gou et~al.(2023)Gou, Shao, Gong, Shen, Yang, Duan, and Chen]{gou2023critic}
Zhibin Gou, Zhihong Shao, Yeyun Gong, Yelong Shen, Yujiu Yang, Nan Duan, and Weizhu Chen.
\newblock Critic: Large language models can self-correct with tool-interactive critiquing.
\newblock \emph{arXiv preprint arXiv:2305.11738}, 2023.

\bibitem[Grossberg(1980)]{grossberg1980}
Stephen Grossberg.
\newblock How does a brain build a cognitive code?
\newblock \emph{Psychological Review}, 87:\penalty0 1--51, 1980.

\bibitem[Ha and Schmidhuber(2018)]{ha2018world}
David Ha and J{\"u}rgen Schmidhuber.
\newblock World {M}odels.
\newblock \emph{arXiv preprint arXiv:1803.10122}, 2018.

\bibitem[Hafner et~al.(2023)Hafner, Pasukonis, Ba, and Lillicrap]{hafner2023mastering}
Danijar Hafner, Jurgis Pasukonis, Jimmy Ba, and Timothy Lillicrap.
\newblock Mastering diverse domains through {W}orld {M}odels.
\newblock \emph{arXiv preprint arXiv:2301.04104}, 2023.

\bibitem[Hansen et~al.(2022)Hansen, Lin, Su, Wang, Kumar, and Rajeswaran]{hansen2022modem}
Nicklas Hansen, Yixin Lin, Hao Su, Xiaolong Wang, Vikash Kumar, and Aravind Rajeswaran.
\newblock Mo{D}em: Accelerating visual model-based reinforcement learning with demonstrations.
\newblock \emph{arXiv preprint arXiv:2212.05698}, 2022.

\bibitem[Hansen et~al.(2023)Hansen, Su, and Wang]{hansen2023td}
Nicklas Hansen, Hao Su, and Xiaolong Wang.
\newblock {TD-MPC}2: Scalable, robust world models for continuous control.
\newblock \emph{arXiv preprint arXiv:2310.16828}, 2023.

\bibitem[Hu et~al.(2021)Hu, Shen, Wallis, Allen-Zhu, Li, Wang, Wang, and Chen]{hu2021lora}
Edward~J Hu, Yelong Shen, Phillip Wallis, Zeyuan Allen-Zhu, Yuanzhi Li, Shean Wang, Lu~Wang, and Weizhu Chen.
\newblock Lo{RA}: Low-rank adaptation of large language models.
\newblock \emph{arXiv preprint arXiv:2106.09685}, 2021.

\bibitem[Hu et~al.(2023)Hu, Tang, Zuo, Wang, Song, Lou, Jiao, and Charles]{hu2023evoke}
Xinyu Hu, Pengfei Tang, Simiao Zuo, Zihan Wang, Bowen Song, Qiang Lou, Jian Jiao, and Denis Charles.
\newblock Evoke: Evoking critical thinking abilities in {LLM}s via reviewer-author prompt editing.
\newblock \emph{arXiv preprint arXiv:2310.13855}, 2023.

\bibitem[Huang et~al.(2023)Huang, Chen, Mishra, Zheng, Yu, Song, and Zhou]{huang2023large}
Jie Huang, Xinyun Chen, Swaroop Mishra, Huaixiu~Steven Zheng, Adams~Wei Yu, Xinying Song, and Denny Zhou.
\newblock Large language models cannot self-correct reasoning yet.
\newblock \emph{arXiv preprint arXiv:2310.01798}, 2023.

\bibitem[Isaacson and Fujita(2006)]{isaacson2006metacognitive}
Randy Isaacson and Frank Fujita.
\newblock Metacognitive knowledge monitoring and self-regulated learning.
\newblock \emph{Journal of the Scholarship of Teaching and Learning}, 6(1): 39--55, 2006.

\bibitem[Jiang et~al.(2023)Jiang, Sablayrolles, Mensch, Bamford, Chaplot, Casas, Bressand, Lengyel, Lample, Saulnier, et~al.]{jiang2023mistral}
Albert~Q Jiang, Alexandre Sablayrolles, Arthur Mensch, Chris Bamford, Devendra~Singh Chaplot, Diego de~las Casas, Florian Bressand, Gianna Lengyel, Guillaume Lample, Lucile Saulnier, et~al.
\newblock Mistral 7{B}.
\newblock \emph{arXiv preprint arXiv:2310.06825}, 2023.

\bibitem[Ketz and Pilly(2022)]{ketz2022}
Nicholas~A Ketz and Praveen~K Pilly.
\newblock Concept-modulated model-based offline reinforcement learning for rapid generalization.
\newblock \emph{arXiv preprint arXiv:2209.03207}, 2022.

\bibitem[Kolouri et~al.(2020)Kolouri, Ketz, Soltoggio, and Pilly]{kolouri2020}
Soheil Kolouri, Nicholas Ketz, Andrea Soltoggio, and Praveen Pilly.
\newblock Sliced {C}ramer synaptic consolidation for preserving deeply learned representations.
\newblock \emph{International Conference on Learning Representations}, 2020.

\bibitem[Koriat(1997)]{koriat1997monitoring}
Asher Koriat.
\newblock Monitoring one's own knowledge during study: A cue-utilization approach to judgments of learning.
\newblock \emph{Journal of Experimental Psychology: General}, 126\penalty0 (4):\penalty0 349, 1997.

\bibitem[Koul et~al.(2020)Koul, Kumar, Fern, and Majumdar]{koul2020dream}
Anurag Koul, Varun~V. Kumar, Alan Fern, and Somdeb Majumdar.
\newblock Dream and search to control: Latent space planning for continuous control.
\newblock \emph{arXiv preprint arXiv:2010.09832}, 2020.

\bibitem[Kramarski and Mevarech(2003)]{kramarski2003}
Bracha Kramarski and Zemira Mevarech.
\newblock Enhancing mathematical reasoning in the classroom: Effects of cooperative learning and metacognitive training.
\newblock \emph{American Educational Research Journal}, 40, 281--310, 2003.

\bibitem[Kruger and Dunning(1999)]{kruger1999}
Justin Kruger and David Dunning.
\newblock Unskilled and unaware of it: How difficulties in recognizing one's own incompetence lead to inflated self-assessments.
\newblock \emph{Journal of Personality and Social Psychology}, 77\penalty0 (6):\penalty0 1121, 1999.

\bibitem[Kudithipudi et~al.(2023)Kudithipudi, Aguilar-Simon, Babb, et~al.]{kudithipudi2023}
Dhireesha Kudithipudi, Mario Aguilar-Simon, Jonathan Babb, et~al.
\newblock Biological underpinnings for lifelong learning machines.
\newblock \emph{Nature Machine Intelligence}, 4(3):\penalty0 196--210, 2023.

\bibitem[Lu et~al.(2025)]{lu2025}
Xiaping Lu, Carsten Murawski, Peter Bossaerts, and Shinsuke Suzuki.
\newblock Estimating self-performance when making complex decisions.
\newblock \emph{Scientific Reports}, 15\penalty0 (1), 3203, 2025.

\bibitem[Madaan et~al.(2023)Madaan, Tandon, Gupta, Hallinan, Gao, Wiegreffe, Alon, Dziri, Prabhumoye, Yang, et~al.]{madaan2023self}
Aman Madaan, Niket Tandon, Prakhar Gupta, Skyler Hallinan, Luyu Gao, Sarah Wiegreffe, Uri Alon, Nouha Dziri, Shrimai Prabhumoye, Yiming Yang, et~al.
\newblock Self-{R}efine: Iterative refinement with self-feedback.
\newblock \emph{arXiv preprint arXiv:2303.17651}, 2023.

\bibitem[Mandi et~al.(2023)Mandi, Bharadhwaj, Moens, Song, Rajeswaran, and Kumar]{mandi2023cacti}
Zhao Mandi, Homanga Bharadhwaj, Vincent Moens, Shuran Song, Aravind Rajeswaran, and Vikash Kumar.
\newblock {CACTI}: A framework for scalable multi-task multi-scene visual imitation learning.
\newblock \emph{arXiv preprint arXiv:2212.05711}, 2023.

\bibitem[Maniscalco and Lau(2012)]{maniscalco2012}
Brian Maniscalco and Hakwan Lau.
\newblock A signal detection theoretic approach for estimating metacognitive sensitivity from confidence ratings.
\newblock \emph{Consciousness and Cognition}, 21\penalty0 (1), 422--430, 2025.

\bibitem[Mehta et~al.(2024)Mehta, Sekhavat, Cao, Horton, Jin, Sun, Mirzadeh, Najibi, Belenko, Zatloukal, et~al.]{mehta2024openelm}
Sachin Mehta, Mohammad~Hossein Sekhavat, Qingqing Cao, Maxwell Horton, Yanzi Jin, Chenfan Sun, Iman Mirzadeh, Mahyar Najibi, Dmitry Belenko, Peter Zatloukal, et~al.
\newblock Open{ELM}: An efficient language model family with open-source training and inference framework.
\newblock \emph{arXiv preprint arXiv:2404.14619}, 2024.

\bibitem[Metcalfe et~al.(1993)Metcalfe, Schwartz, and Joaquim]{metcalfe1993cue}
Janet Metcalfe, Bennett~L Schwartz, and Scott~G Joaquim.
\newblock The cue-familiarity heuristic in metacognition.
\newblock \emph{Journal of Experimental Psychology: Learning, Memory, and Cognition}, 19\penalty0 (4):\penalty0 851, 1993.

\bibitem[Micheli et~al.(2023)Micheli, Alonso, and Fleuret]{micheli2023transformers}
Vincent Micheli, Eloi Alonso, and François Fleuret.
\newblock Transformers are sample-efficient {W}orld {M}odels.
\newblock \emph{arXiv preprint arXiv:2209.00588}, 2023.

\bibitem[Middlebrooks et~al.(2012)Middlebrooks and Sommer]{middlebrooks2012}
Paul Middlebrooks and Marc Sommer.
\newblock Neuronal correlates of metacognition in primate frontal cortex.
\newblock \emph{Neuron}, 75\penalty0 (3):\penalty0 517--530, 2012.

\bibitem[Miyamoto et~al.(2017)Miyamoto, Osada, Setsuie, Takeda, Tamura, Adachi, and Miyashita]{miyamoto2017causal}
Kentaro Miyamoto, Takahiro Osada, Rieko Setsuie, Masaki Takeda, Keita Tamura, Yusuke Adachi, and Yasushi Miyashita.
\newblock Causal neural network of metamemory for retrospection in primates.
\newblock \emph{Science}, 355\penalty0 (6321):\penalty0 188--193, 2017.

\bibitem[Moerland et~al.(2023)Moerland, Broekens, Plaat, Jonker]{moerland2023}
Thomas Moerland, Joost Broekens, Aske Plaat, and Catholijn Jonker.
\newblock Model-based reinforcement learning: A survey.
\newblock \emph{Foundations and Trends® in Machine Learning}, 16\penalty0 (1):\penalty0 1--118, 2023.

\bibitem[Mukherjee et~al.(2023)Mukherjee, Mitra, Jawahar, Agarwal, Palangi, and Awadallah]{mukherjee2023orca}
Subhabrata Mukherjee, Arindam Mitra, Ganesh Jawahar, Sahaj Agarwal, Hamid Palangi, and Ahmed Awadallah.
\newblock Orca: Progressive learning from complex explanation traces of {GPT}-4.
\newblock \emph{arXiv preprint arXiv:2306.02707}, 2023.

\bibitem[Nelson and Narens(1990)]{nelsonnarens1990}
Thomas~O Nelson and Louis Narens.
\newblock \emph{Metamemory: A theoretical framework and new findings}.
\newblock Academic Press, 1990.

\bibitem[Open-Orca(2024)]{OpenOrcaMistral7B}
Open-Orca.
\newblock Mistral-7{B}-{O}pen{O}rca.
\newblock \url{https://huggingface.co/Open-Orca/Mistral-7B-OpenOrca}, 2024.

\bibitem[Paul et~al.(2023)Paul, Ismayilzada, Peyrard, Borges, Bosselut, West, and Faltings]{paul2023refiner}
Debjit Paul, Mete Ismayilzada, Maxime Peyrard, Beatriz Borges, Antoine Bosselut, Robert West, and Boi Faltings.
\newblock Refiner: Reasoning feedback on intermediate representations.
\newblock \emph{arXiv preprint arXiv:2304.01904}, 2023.

\bibitem[Rafailov et~al.(2024)Rafailov, Sharma, Mitchell, Manning, Ermon, and Finn]{rafailov2024}
Rafael Rafailov, Archit Sharma, Eric Mitchell, Christopher~D. Manning, Stefano Ermon, and Chelsea Finn.
\newblock Direct {P}reference {O}ptimization: Your language model is secretly a reward model.
\newblock \emph{Advances in Neural Information Processing Systems}, 36, 2024.

\bibitem[Robine et~al.(2023)Robine, H{\"o}ftmann, Uelwer, and Harmeling]{robine2023transformer}
Jan Robine, Marc H{\"o}ftmann, Tobias Uelwer, and Stefan Harmeling.
\newblock Transformer-based {W}orld {M}odels are happy with 100{K} interactions.
\newblock \emph{arXiv preprint arXiv:2303.07109}, 2023.

\bibitem[Rostami et~al.(2019)Rostami, Kolouri, and Pilly]{rostami2019}
Mohammad Rostami, Soheil Kolouri, and Praveen Pilly.
\newblock Complementary learning for overcoming catastrophic forgetting using experience replay.
\newblock \emph{arXiv preprint arXiv:1903.04566}, 2019.

\bibitem[Schraw(1998)]{schraw1998promoting}
Gregory Schraw.
\newblock Promoting general metacognitive awareness.
\newblock \emph{Instructional Science}, 26:\penalty0 113--125, 1998.

\bibitem[Schraw et~al.(2006)Schraw, Crippen, and Hartley]{schraw2006}
Gregory Schraw, Kent Crippen, and Kendall Hartley.
\newblock Promoting self-regulation in science education: Metacognition as part of a broader perspective on learning.
\newblock \emph{Research in Science Education}, 36, 111--139, 2006.

\bibitem[Sentence{T}ransformers(2024)]{SentenceTransformers}
Sentence{T}ransformers.
\newblock all-{M}ini{LM}-{L6}-v2.
\newblock \url{https://huggingface.co/sentence-transformers/all-MiniLM-L6-v2}, 2024.

\bibitem[Shinn et~al.(2023)Shinn, Cassano, Gopinath, Narasimhan, and Yao]{shinn2023reflexion}
Noah Shinn, Federico Cassano, Ashwin Gopinath, Karthik Narasimhan, and Shunyu Yao.
\newblock Reflexion: Language agents with verbal reinforcement learning.
\newblock \emph{Advances in Neural Information Processing Systems}, 36:\penalty0 8634--8652, 2023.

\bibitem[Shridhar et~al.(2020)Shridhar, Yuan, C{\^o}t{\'e}, Bisk, Trischler, and Hausknecht]{shridhar2020alfworld}
Mohit Shridhar, Xingdi Yuan, Marc-Alexandre C{\^o}t{\'e}, Yonatan Bisk, Adam Trischler, and Matthew Hausknecht.
\newblock {ALFW}orld: Aligning text and embodied environments for interactive learning.
\newblock \emph{arXiv preprint arXiv:2010.03768}, 2020.

\bibitem[Silver et~al.(2016)Silver, Huang, Maddison, Guez, Sifre, Van Den~Driessche, Schrittwieser, Antonoglou, Panneershelvam, Lanctot, et~al.]{silver2016mastering}
David Silver, Aja Huang, Chris~J Maddison, Arthur Guez, Laurent Sifre, George Van Den~Driessche, Julian Schrittwieser, Ioannis Antonoglou, Veda Panneershelvam, Marc Lanctot, et~al.
\newblock Mastering the game of {G}o with deep neural networks and tree search.
\newblock \emph{Nature}, 529\penalty0 (7587):\penalty0 484--489, 2016.

\bibitem[Silver et~al.(2017)Silver, Schrittwieser, Simonyan, Antonoglou, Huang, Guez et~al.]{silver2017}
David Silver, Julian Schrittwieser, Karen Simonyan, et~al.
\newblock Mastering the game of {G}o without human knowledge.
\newblock \emph{Nature}, 550\penalty0 (7676):\penalty0 354--359, 2017.

\bibitem[Silver et~al.(2018)Silver, Hubert, Schrittwieser, Antonoglou, Lai, Guez et~al.]{silver2018}
David Silver, Thomas Hubert, Julian Schrittwieser, et~al.
\newblock A general reinforcement learning algorithm that masters chess, shogi, and {G}o through self-play.
\newblock \emph{Science}, 362\penalty0 (6419):\penalty0 1140--1144, 2018.

\bibitem[Sutton(1991)]{sutton1991}
Richard~S. Sutton.
\newblock Dyna, an integrated architecture for learning, planning, and reacting.
\newblock \emph{SIGART Bulletin}, 2\penalty0 (4):\penalty0 160–163, 1991.

\bibitem[Team et~al.(2023)Team, Bauer, Baumli, Baveja, Behbahani, Bhoopchand, Bradley-Schmieg, Chang, Clay, Collister, Dasagi, Gonzalez, Gregor, Hughes, Kashem, Loks-Thompson, Openshaw, Parker-Holder, Pathak, Perez-Nieves, Rakicevic, Rocktäschel, Schroecker, Sygnowski, Tuyls, York, Zacherl, and Zhang]{adaptiveagentteam2023humantimescale}
Adaptive~Agent Team, Jakob Bauer, Kate Baumli, Satinder Baveja, Feryal Behbahani, Avishkar Bhoopchand, Nathalie Bradley-Schmieg, Michael Chang, Natalie Clay, Adrian Collister, Vibhavari Dasagi, Lucy Gonzalez, Karol Gregor, Edward Hughes, Sheleem Kashem, Maria Loks-Thompson, Hannah Openshaw, Jack Parker-Holder, Shreya Pathak, Nicolas Perez-Nieves, Nemanja Rakicevic, Tim Rocktäschel, Yannick Schroecker, Jakub Sygnowski, Karl Tuyls, Sarah York, Alexander Zacherl, and Lei Zhang.
\newblock Human-timescale adaptation in an open-ended task space.
\newblock \emph{arXiv preprint arXiv:2301.07608}, 2023.

\bibitem[Team et~al.(2021)Team, Stooke, Mahajan, Barros, Deck, Bauer, Sygnowski, Trebacz, Jaderberg, Mathieu, McAleese, Bradley-Schmieg, Wong, Porcel, Raileanu, Hughes-Fitt, Dalibard, and Czarnecki]{openendedlearningteam2021openended}
Open Ended~Learning Team, Adam Stooke, Anuj Mahajan, Catarina Barros, Charlie Deck, Jakob Bauer, Jakub Sygnowski, Maja Trebacz, Max Jaderberg, Michael Mathieu, Nat McAleese, Nathalie Bradley-Schmieg, Nathaniel Wong, Nicolas Porcel, Roberta Raileanu, Steph Hughes-Fitt, Valentin Dalibard, and Wojciech~Marian Czarnecki.
\newblock Open-ended learning leads to generally capable agents.
\newblock \emph{arXiv preprint arXiv:2107.12808}, 2021.

\bibitem[Wang et~al.(2021)Wang, Zhao, and Li]{wang2021}
Haoxiang Wang, Han Zhao, and Bo~Li.
\newblock Bridging multi-task learning and meta-learning: Towards efficient training and effective adaptation.
\newblock \emph{International Conference on Machine Learning}, 10991--11002, 2021.

\bibitem[Wang et~al.(2022)Wang, Wei, Schuurmans, Le, Chi, Narang, Chowdhery, and Zhou]{wang2022self}
Xuezhi Wang, Jason Wei, Dale Schuurmans, Quoc Le, Ed~Chi, Sharan Narang, Aakanksha Chowdhery, and Denny Zhou.
\newblock Self-consistency improves chain of thought reasoning in language models.
\newblock \emph{arXiv preprint arXiv:2203.11171}, 2022.

\bibitem[Webb et~al.(2023)Webb, Miyoshi, So, Rajananda, and Lau]{webb2023}
Taylor Webb, Kiyofumi Miyoshi, Tsz Yan So, Sivananda Rajananda, and Hakwan Lau.
\newblock Natural statistics support a rational account of confidence biases.
\newblock \emph{Nature Communications}, 14(1), 10991--11002, 2023.

\bibitem[Wei et~al.(2022)Wei, Wang, Schuurmans, Bosma, Chi, Le, and Zhou]{wei2022chain}
Jason Wei, Xuezhi Wang, Dale Schuurmans, Maarten Bosma, Ed~Chi, Quoc Le, and Denny Zhou.
\newblock Chain-of-thought prompting elicits reasoning in large language models.
\newblock \emph{arXiv preprint arXiv:2201.11903}, 2022.

\bibitem[Xie et~al.(2023)Xie, Kawaguchi, Zhao, Zhao, Kan, He, and Xie]{xie2023decomposition}
Yuxi Xie, Kenji Kawaguchi, Yiran Zhao, Xu~Zhao, Min-Yen Kan, Junxian He, and Qizhe Xie.
\newblock Decomposition enhances reasoning via self-evaluation guided decoding.
\newblock \emph{arXiv preprint arXiv:2305.00633}, 2023.

\bibitem[Yao et~al.(2022)Yao, Zhao, Yu, Du, Shafran, Narasimhan, and Cao]{yao2023react}
Shunyu Yao, Jeffrey Zhao, Dian Yu, Nan Du, Izhak Shafran, Karthik Narasimhan, and Yuan Cao.
\newblock {ReAct}: Synergizing reasoning and acting in language models.
\newblock \emph{arXiv preprint arXiv:2210.03629}, 2022.

\bibitem[Yao et~al.(2024)Yao, Yu, Zhao, Shafran, Griffiths, Cao, and Narasimhan]{yao2024tree}
Shunyu Yao, Dian Yu, Jeffrey Zhao, Izhak Shafran, Tom Griffiths, Yuan Cao, and Karthik Narasimhan.
\newblock Tree of thoughts: Deliberate problem solving with large language models.
\newblock \emph{Advances in Neural Information Processing Systems}, 36, 2024.

\bibitem[Young and Fry(2008)]{young2008metacognitive}
Andria Young and Jane~D Fry.
\newblock Metacognitive awareness and academic achievement in college students.
\newblock \emph{Journal of the Scholarship of Teaching and Learning}, 8\penalty0 (2):\penalty0 1--10, 2008.

\bibitem[Yu et~al.(2020)Yu, Quillen, He, Julian, Hausman, Finn, and Levine]{yu2020meta}
Tianhe Yu, Deirdre Quillen, Zhanpeng He, Ryan Julian, Karol Hausman, Chelsea Finn, and Sergey Levine.
\newblock Meta-{W}orld: A benchmark and evaluation for multi-task and meta reinforcement learning.
\newblock \emph{Conference on Robot Learning}, 1094--1100, 2020.

\bibitem[Zhou et~al.(2022)Zhou, Schärli, Hou, Wei, Scales, Wang, Schuurmans, Bousquet, Le, and Chi]{zhou2022least}
Denny Zhou, Nathanael Schärli, Le~Hou, Jason Wei, Nathan Scales, Xuezhi Wang, Dale Schuurmans, Olivier Bousquet, Quoc Le, and Ed~Chi.
\newblock Least-to-most prompting enables complex reasoning in large language models.
\newblock \emph{arXiv preprint arXiv:2205.10625}, 2022.

\end{thebibliography}
